\documentclass{article}

\usepackage{arxiv}

\usepackage[utf8]{inputenc} 
\usepackage[T1]{fontenc}    
\usepackage{hyperref}       
\usepackage{url}            
\usepackage{booktabs}       
\usepackage{amsfonts}       
\usepackage{nicefrac}       
\usepackage{microtype}      
\usepackage{lipsum}		
\usepackage{graphicx}
\usepackage{doi}
\usepackage{multirow}
\usepackage{caption}
\usepackage{subcaption}
\usepackage{booktabs}
\usepackage{authblk}
\usepackage{setspace}
\usepackage{array}
\usepackage{todonotes}
\usepackage{longtable}

\usepackage[sort, numbers, square]{natbib}

\usepackage[acronym,nohypertypes={acronym, notation}]{glossaries}
\makeglossaries

\newacronym{MCTS}{MCTS}{Monte-Carlo Tree Search}
\newacronym{DFPN}{DFPN}{Depth-First Proof-Number}
\newacronym{AZF}{AZF}{AiZynthFinder}

\newacronym[shortplural={RNNs}, longplural={Recurrent Neural Networks}]{RNN}{RNN}{Recurrent Neural Network}

\title{Models Matter: The Impact of Single-Step Retrosynthesis on Synthesis Planning}

\date{} 					

\author[1, 2]{\underline{Paula Torren-Peraire}$^*$}
\author[3, 4]{\underline{Alan Kai Hassen}$^*$}
\author[5]{Samuel Genheden}
\author[2]{Jonas Verhoeven}
\author[4]{Djork-Arné Clevert}
\author[1]{Mike Preuss}
\author[2]{Igor Tetko}

\affil[1]{\textsl{\small Institute of Structural Biology\\
  Molecular Targets and Therapeutics Center\\
  Helmholtz Zentrum München\\
  Neuherberg, Germany}}
\affil[2]{\textsl{\small In-Silico Discovery and External Innovation\\
  Janssen Research \& Development\\
  Janssen Pharmaceutica N.V\\ 
  Beerse, Belgium}}
\affil[3]{\textsl{\small Leiden Institute of Advanced Computer Science\\
  Leiden University\\
  Leiden, The Netherlands}}
\affil[4]{\textsl{\small Machine Learning Research\\
    Pfizer Worldwide Research Development and Medical\\ 
    Berlin, Germany}}
\affil[5]{\textsl{\small Molecular AI\\ 
  Discovery Sciences, R\&D\\
  AstraZeneca\\
  Gothenburg, Sweden}}

\affil[*]{\textbf{\small The following authors contributed equally.}}

\renewcommand{\headeright}{}
\renewcommand{\undertitle}{}
\renewcommand{\shorttitle}{Models Matter: The Impact of Single-Step Retrosynthesis on Synthesis Planning}

\hypersetup{
pdftitle={Models Matter: The Impact of Single-Step Retrosynthesis on Synthesis Planning},
pdfauthor={Paula Torren-Peraire, Alan Kai Hassen, Samuel Genheden, Jonas Verhoeven, Djork-Arne Clevert, Mike Preuss, Igor Tetko},
pdfkeywords={Computer-Aided Synthesis Planning, Retrosynthesis, Synthesis, Retrosynthesis Prediction, Benchmark, Retro*, LocalRetro, MHNreact, Chemformer, Template-Based, AiZynthFinder},
}

\begin{document}
\onehalfspacing
\maketitle

\begin{abstract}
Retrosynthesis consists of breaking down a chemical compound recursively step-by-step into molecular precursors until a set of commercially available molecules is found with the goal to provide a synthesis route. Its two primary research directions, single-step retrosynthesis prediction, which models the chemical reaction logic, and multi-step synthesis planning, which tries to find the correct sequence of reactions, are inherently intertwined. Still, this connection is not reflected in contemporary research. In this work, we combine these two major research directions by applying multiple single-step retrosynthesis models within multi-step synthesis planning and analyzing their impact using public and proprietary reaction data. We find a disconnection between high single-step performance and potential route-finding success, suggesting that single-step models must be evaluated within synthesis planning in the future. Furthermore, we show that the commonly used single-step retrosynthesis benchmark dataset USPTO-50k is insufficient as this evaluation task does not represent model performance and scalability on larger and more diverse datasets. For multi-step synthesis planning, we show that the choice of the single-step model can improve the overall success rate of synthesis planning by up to +28\% compared to the commonly used baseline model. Finally, we show that each single-step model finds unique synthesis routes, and differs in aspects such as route-finding success, the number of found synthesis routes, and chemical validity, making the combination of single-step retrosynthesis prediction and multi-step synthesis planning a crucial aspect when developing future methods. 
\end{abstract}



\section{Introduction}
The Design-Make-Test-Analyse (DMTA) cycle is commonly used in small molecule drug discovery to explore novel compounds and indications. Over recent years, it has seen massive changes with the introduction of modern machine-learning approaches \cite{vijayanEnhancingPreclinicalDrug2022}. Retrosynthesis, a core task in the Make part of the DMTA cycle of modern drug discovery, is a technique commonly used by organic chemists in synthesis planning. A molecule is successively broken down into smaller subunits until easily synthesizable or purchasable compounds are obtained \cite{seglerPlanningChemicalSyntheses2018, coreyLogicChemicalSynthesis1989}, where the overall goal is to produce a roadmap for the synthesis of a target compound. With computer-aided retrosynthesis, researchers in both chemistry and machine learning aim to accelerate the development of chemical synthesis by saving time and resources, addressing more complex molecules or producing more efficient and safe routes. These generated routes can be used by medical chemists to create molecules of interest \cite{coleyMachineLearningComputerAided2018}, serve as a basis for autonomous chemistry \cite{coleyRoboticPlatformFlow2019}, or be incorporated into De Novo Drug Design to assess synthesizability \cite{miljkovicImpactArtificialIntelligence2021}.

The core advance in retrosynthesis has been the realignment with common machine learning approaches \cite{schwallerMachineIntelligenceChemical2022} which allow users to consider a much larger set of potential synthesis routes. The machine learning field of retrosynthesis prediction is commonly separated into two research fields, referred to as single-step retrosynthesis prediction and multi-step synthesis planning. Where single-step retrosynthesis prediction refers to breaking down a product into a single set of reactants and multi-step synthesis planning refers to the search algorithms used to find synthesis routes leading to purchasable compounds (building blocks). 

Specifically, single-step retrosynthesis prediction is a supervised learning task, developed to predict which reactions are relevant to a target molecule, and the corresponding reactants required to produce this reaction. There are two commonly referenced categories of single-step retrosynthesis models, template-based and template-free \cite{schwallerMachineIntelligenceChemical2022}. 
Template-based methods use reaction templates, an abstraction of the reactions in the data, which summarize the underlying pattern of these reactions. There are different approaches to extracting templates, though in all cases these processes aim to represent the atom and bond structures required to perform a reaction \cite{zhongRecentAdvancesArtificial2023}, where a single template will represent multiple reactions. Template-based methods consider single-step prediction as a classification problem where the task is to predict the appropriate template for the target molecule/product. Examples of template-based methods include NeuralSym \cite{seglerNeuralSymbolicMachineLearning2017}, the first approach in the field which demonstrated the usefulness of using deep neural networks for retrosynthesis prediction, MHNreact \cite{seidlImprovingFewZeroShot2022}, which uses an information retrieval approach to associate products and templates and LocalRetro \cite{chenDeepRetrosyntheticReaction2021}, which uses a graph representation to predict relevant local atom and bond templates for the product. 

On the other hand, template-free approaches commonly treat retrosynthesis prediction as a sequence-to-sequence prediction problem \cite{zhongRecentAdvancesArtificial2023}, employing methods seen in natural language processing such as language translation tasks. Instead of extracting and predicting the corresponding templates, the approach aims to learn the underlying reactions to directly predict reactants. Product and reactants are typically introduced as Simplified Molecular-Input Line-Entry System (SMILES), a common text-based representation of chemical entities. Examples of template-free methods include Chemformer \cite{irwinChemformerPretrainedTransformer2022}, a large pretrained transformer model fine-tuned on the retrosynthesis task, and Augmented Transformer \cite{tetkoStateoftheartAugmentedNLP2020}, a transformer architecture which employs multiple types of augmentation. Other variations of these approaches exist, such as semi-template-based where the molecule is first broken down into subparts then completed to produce chemically viable reactants \cite{somnathLearningGraphModels2021, shiGraphGraphsFramework2020, wangRetroPrimeDiversePlausible2021}.

Multi-step synthesis planning focuses on researching novel synthesis route search algorithms using a single-step model to identify retrosynthetic disconnections. The pioneering approach in the field uses Monte Carlo Tree Search (MCTS) to plan the traversal of the search tree at run time guided by a neural network \cite{seglerPlanningChemicalSyntheses2018}. Alternative route planning algorithms use an oracle function or heuristics to guide the tree search instead of relying on compute-expensive run time planning. Prominent examples of this are Depth-First Proof-Number (DFPN) \cite{kishimotoDepthFirstProofNumberSearch2019}, which combines classical DFPN with a neural heuristic, Retro*, which combines A* pathfinding with a neural heuristic \cite{chenRetroLearningRetrosynthetic2020}, or RetroGraph, which applies a holistic graph-based approach \cite{xieRetroGraphRetrosyntheticPlanning2022}. Other approaches incorporate reaction feasibility into the tree search \cite{linAutomaticRetrosyntheticRoute2020} or use synthesizability heuristics in combination with a forward synthesis model \cite{schwallerPredictingRetrosyntheticPathways2020, kreutterMultistepRetrosynthesisCombining2023}. Finally, self-play approaches, motivated by their success in Go \cite{silverMasteringGameGo2017}, learn to guide the tree search by leveraging information gathered from prior runs of synthesis planning \cite{schreckLearningRetrosyntheticPlanning2019, yuGRASPNavigatingRetrosynthetic2022, liuRetrosyntheticPlanningDual2023}. 

Single-step retrosynthesis prediction and multi-step synthesis planning are inherently intertwined where the single-step method defines the maximum searchable reaction network, and the search algorithm tries to efficiently traverse this network by repeatedly applying the chemical information that is stored in the single-step model. However, this connection is not reflected in contemporary research. 

Currently, single-step methods are benchmarked by predicting a single retrosynthetic step from a product to reactants. The common benchmark data for these methods, USPTO-50k \cite{loweExtractionChemicalStructures2012, schneiderWhatWhatNearly2016}, consists of around 50k reactions and only has a limited diversity of 10 reaction classes. These methods are typically only tested on reactant prediction and not within multi-step search algorithms, therefore their usability for synthesis planning is not assessed. Similarly, multi-step search algorithms benchmark the route-finding capabilities of their method using a single single-step model, often based on the template-based NeuralSym model \cite{seglerPlanningChemicalSyntheses2018, liuRetrosyntheticPlanningDual2023, xieRetroGraphRetrosyntheticPlanning2022, kishimotoDepthFirstProofNumberSearch2019, chenRetroLearningRetrosynthetic2020}, and evaluate the success rate of finding potential synthesis routes for molecules of interest. However, the approach of using only one single-step model does not consider the impact of alternative single-step models, a vital aspect of the search, as the route planning algorithm uses the reaction information stored in the single-step model to find synthesis routes and create alternate reaction pathways within the reaction network.

The current question remains whether state-of-the-art single-step retrosynthesis methods are transferable to the multi-step synthesis planning domain, and their impact on multi-step synthesis planning \cite{tuRetrosynthesisPredictionRevisited2022, hassenMindRetrosynthesisGap2022}. In this work, we address the transfer between single-step and multi-step methods by incorporating different state-of-the-art single-step models within a common multi-step search algorithm to analyze the use of these models for multi-step synthesis planning. We explore the effect on performance, analyzing the relationship between contemporary single-step and multi-step performance metrics using both public and proprietary datasets of varying size and diversity. Moreover, we also focus on vital aspects such as model suitability and chemical validity of the predicted routes.

\section{Methods}

\begin{figure}[htb]
    \normalsize
    \centering
    \includegraphics[width=0.7\textwidth]{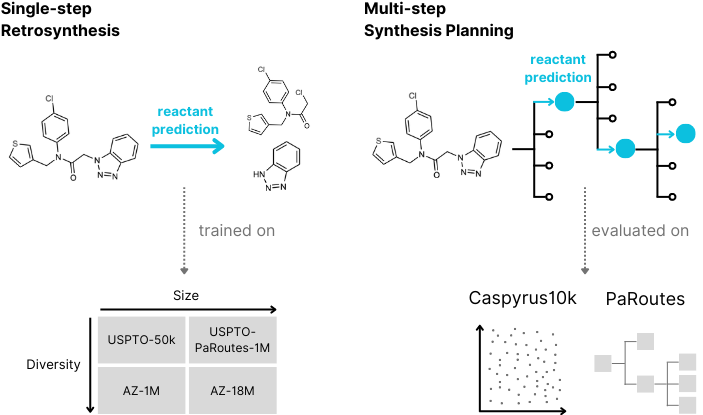} 
    \caption{Evaluation Framework for single-step models (AiZynthFinder (AZF), LocalRetro, Chemformer, and MHNreact), trained on different public (USPTO-50k, USPTO-PaRoutes-1M) and proprietary (AZ-1M, AZ-18M) datasets in synthesis planning on Caspyrus10k and PaRoutes.}
    \label{fig:graphic_abstract}
\end{figure}

In this work, we develop an evaluation framework to benchmark different single-step models in multi-step synthesis planning (Figure \ref{fig:graphic_abstract}).

\subsection{Evaluation Scheme}
\begin{table}[h]
        \centering
        \caption{Evaluation metrics for single-step retrosynthesis and multi-step synthesis planning. Solved synthesis route implies that the produced route leads to building blocks. }
        \label{tab:evaluation_scheme}

        \begin{tabular}{@{}m{8em} m{10em} m{19em}@{}}
        \toprule
\textbf{Task} & \textbf{Metric} & \textbf{Description} \\ \midrule
\begin{tabular}[r]{@{}l@{}}Single-Step\\Retrosynthesis\end{tabular}    & Top-N Accuracy           & Percentage of compounds for which the ground-truth reactants are predicted within the top-N \\ \hline
\multirow{2}{*}{\begin{tabular}[r]{@{}l@{}}Multi-Step \\ Synthesis Planning\end{tabular}} & Success Rate &
Percentage of compounds where at least one solved synthesis route is produced \\ \cline{2-3} 
                              & Number of Solved Routes & Average number of unique solved synthesis routes produced per molecule      \\ \cline{2-3}
                                & Search Times & Average search time per molecule \\ \cline{2-3}
                                & Single-Step Model Calls & Average number of single-step model calls per molecule \\ \cline{2-3} 
                                & Route Accuracy & Percentage of compounds where the gold-standard route is predicted within the top-N synthesis routes \\ \cline{2-3}
                                & Building Block Accuracy & Percentage of compounds where the gold-standard building blocks are predicted within the top-N synthesis routes \\ 
    
        \bottomrule
       \end{tabular}
       
\end{table}

\textbf{Single-step Retrosynthesis.} Single-step retrosynthesis methods are evaluated using top-n accuracy \cite{schwallerMachineIntelligenceChemical2022} (Table \ref{tab:evaluation_scheme}). The task for single-step retrosynthesis is the correct prediction of (gold-standard) reactants from the product of a known reaction. Here, we measure the percentage of target molecules for which the correct reactants are recovered within top-n predictions. Considering that the single-step model defines a possible maximum reaction network for a molecule of interest, published reactions are used to assess the accuracy of the single-step model since they are assumed to be chemically valid. Consequently, the assumption is that if the single-step model can recover a greater number of published reactants, then the predictions produced by the model are chemically viable reactions.

\textbf{Multi-step Synthesis Planning.} On the other hand, for multi-step synthesis planning, the task is the search for likely synthesis routes for a molecule of interest, i.e., a reaction pathway from the target molecule to a set of available building blocks \cite{schwallerMachineIntelligenceChemical2022}. For this, we consider multiple aspects for both the search and the predicted routes.
 
Within success rate, we measure the percentage of molecules for which the route planning algorithm can successfully return at least one solved synthesis route leading from a molecule to building blocks. This condition is required for synthesis routes since a chemist can only consider routes as a suggestion for experimental evaluation if a complete synthesis route is found. Moreover, we analyze the number of solved routes since not only is it interesting to identify if there is a possible synthesis route for a molecule, but also how many alternatives are produced, given that different synthesis routes have different route properties.
 
Nevertheless, algorithmic success does not measure if a found synthesis route is chemically valid, but only if a route into building blocks is found. Route accuracy is used to measure the chemical validity of synthesis routes as predicted routes can be compared to published, experimentally tested gold-standard routes \cite{genhedenPaRoutesFrameworkBenchmarking2022}. Naturally, a route planning algorithm should be able to recover the gold-standard routes within the set of predicted, solved synthesis routes. This task is inherently more complex than producing solved routes (success rate) since it requires a sequence of multiple reactions and their intermediates to be correctly predicted and in the correct order. Additionally, we calculate whether there is an exact match between the predicted building blocks and the gold-standard building blocks. Building block accuracy differs from route accuracy since the route reactions and intermediates are not considered. In all cases it must be noted that a gold standard route is only one possible way of synthesizing a target molecule. 
 
Lastly, we consider search times and single-step model calls. Ideally, synthesis planning algorithms should produce routes in a timely manner to reduce allocated computational resources. However, different single-step models can have different inference speeds, and the time required for a search can massively diverge \cite{hassenMindRetrosynthesisGap2022}. Consequently, the average search time for a molecule with a fixed number of single-step model calls, is measured. Additionally, we report the number of single-step model calls since, in some cases, the method may not reach the maximum iteration limit in the maximum search time. Noteworthy, the maximum search time can be exceeded if the last search iteration is started before the time limit is reached.

\subsection{Datasets.} 

\begin{table}[h]
        \centering
        \caption{Datasets for training single-step retrosynthesis models and evaluating multi-step synthesis planning}
       \label{tab:datasets}

        \begin{tabular}{@{}m{8em} m{10em} m{19em}@{}}
        \toprule
\textbf{Task} & \textbf{Dataset} & \textbf{Description} \\ \midrule
\multirow{2}{*}{\begin{tabular}[r]{@{}l@{}}Single-Step \\ Retrosynthesis \\ Training\end{tabular}}  & USPTO-50k \cite{schneiderWhatWhatNearly2016} & Default single-step retrosynthesis benchmark dataset  \\ \cline{2-3} 
& USPTO-PaRoutes-1M \cite{genhedenPaRoutesFrameworkBenchmarking2022} & Largest publicly available single-step  retrosynthesis dataset \\ \cline{2-3}

& AZ-1M \cite{genhedenAiZynthTrainRobustReproducible2023} & 1M reaction subsample of internal AstraZeneca reactions \\ \cline{2-3}  
& AZ-18M \cite{genhedenAiZynthTrainRobustReproducible2023} & Dataset based on internal AstraZeneca reactions  \\ \hline
\multirow{2}{*}{\begin{tabular}[r]{@{}l@{}}Multi-Step\\Synthesis Planning \\ Evaluation\end{tabular}}  & Caspyrus10k & 10,000 clustered bio-active molecules from Papyrus \cite{bequignonPapyrusLargescaleCurated2023}  \\ \cline{2-3} 
& PaRoutes \cite{genhedenPaRoutesFrameworkBenchmarking2022} & Collection of 10,000 gold-standard synthesis routes extracted from patents \\  

        \bottomrule
       \end{tabular}
\end{table}

\textbf{Single-step Retrosynthesis.} Within single-step retrosynthesis datasets, each reaction is unique. They are all curated to comprise a single product leading to one or more reactants. One product can have more than one recorded reaction, and a reaction type can occur multiple times. Here we use four different single-step retrosynthesis datasets, USPTO-50k \cite{schneiderWhatWhatNearly2016}, USPTO-PaRoutes-1M \cite{genhedenPaRoutesFrameworkBenchmarking2022}, AZ-1M and AZ-18M \cite{genhedenAiZynthTrainRobustReproducible2023} (Table \ref{tab:datasets}). USPTO-50k is the default benchmark dataset for single-step retrosynthesis prediction. It features 50,016 reactions comprising ten reaction classes extracted from the original USPTO dataset \cite{loweExtractionChemicalStructures2012}, which originates from the United States Patent and Trademark Office. USPTO-PaRoutes-1M is a processed version of the original USPTO grant and application data. This single-step dataset is specifically developed to train single-step retrosynthesis models to benchmark multi-step algorithms \cite{genhedenPaRoutesFrameworkBenchmarking2022}. The dataset contains single-step reactions and excludes gold-standard synthesis routes and their corresponding reactions for multi-step benchmarking. Here, we use the PaRoutes 2.0 dataset, which contains 1,198,554 single-step reactions \cite{genhedenAiZynthTrainRobustReproducible2023}. 
 
Additionally, we use two datasets based on the proprietary AstraZeneca dataset \cite{genhedenAiZynthTrainRobustReproducible2023, thakkarDatasetsTheirInfluence2020}. The first, AZ-18M, is the complete cleaned dataset from AstraZeneca, which includes Reaxys \cite{ elsevierlimitedReaxys2023}, Pistachio (a superset of USPTO-PaRoutes-1M) \cite{nextmovesoftwarePistachio2023}, and AstraZeneca Electronic Laboratory Notebooks (ELN) data. This dataset contains 18,697,432 single-step reactions \cite{genhedenAiZynthTrainRobustReproducible2023}. Moreover, to obtain a dataset representative of AZ-18M with a comparable size to USPTO-PaRoutes-1M, we randomly subsample 1M reactions from AZ-18M to produce AZ-1M.
 
To evaluate single-step models, we split all reaction datasets into random 80\% training, 10\% validation, and 10\% test hold-out splits. In the case of USPTO-PaRoutes-1M, to replicate the original data split size \cite{genhedenPaRoutesFrameworkBenchmarking2022}, the hold-out split ratio is 90\% training, 5\% validation, and 5\% test. We defer from using the original hold-out splits since they are based on template stratification. For AZ-18M, we subsample 100k molecules from the complete test set of 1.8 million reactions to avoid excessive evaluation computation.

\textbf{Multi-step Synthesis Planning.} Multi-step evaluation datasets are collections of compounds that are used to test the route-finding capabilities of multi-step synthesis planning algorithms.  To evaluate the synthesis planning capabilities of different single-step models, we create a new dataset called Caspyrus10k that consists of a clustered set of 10,000 molecules from a selection of known bioactive and synthesizable compounds, to ensure a reasonable representation of the synthesizable chemical space.

In detail, we select the high-quality Papyrus \cite{bequignonPapyrusLargescaleCurated2023} dataset of 1,238,835 molecules, where each molecule has an exact bioactivity value measure and is associated with a single protein, strongly suggesting that each of those molecules is synthesizable as its activity has been tested in an experimental setting. We filter those molecules with the Guacamol cleaning strategy \cite{brownGuacaMolBenchmarkingModels2019} to ensure drug-like molecules, removing molecules which do not fit the criteria in the process. As we are interested in these molecules for synthesis planning, we remove the building blocks present in Zinc \cite{genhedenAiZynthFinderFastRobust2020}, Enamine  \cite{enamineltd.EnamineBuildingBlocks2023}, MolPort \cite{molportsiaMoldportCompoundSourcing2023}, and eMolecules \cite{EMoleculesChemicalBuilding2023}. Finally, we cluster the resulting set of molecules using Butina Clustering \cite{butinaUnsupervisedDataBase1999} using Morgan Fingerprints with a radius of 2, a fingerprint size of 1024, and a Butina cut-off threshold of 0.6. From the resulting cluster centroids, we remove 19 centroids in clinical phases 1-3 since they are intellectual property. Finally, we take the largest 10,000 cluster centroids, representing roughly 284,000 molecules.

Additionally, we evaluate the synthesis planning capabilities of all single-step models on PaRoutes \cite{genhedenPaRoutesFrameworkBenchmarking2022}, a collection of 10,000 gold-standard retrosynthesis routes. This task differs from the general synthesis planning task with Caspyrus10k in that the goal is to recover specific real-world synthesis routes conducted as part of a patent application process and therefore test the chemical validity of the predicted synthesis routes. The gold-standard routes are obtained from USPTO patent data, where we use the n-1 set, which contains a single retrosynthesis route for each patent. As stated in the PaRoutes dataset, we use a specialized set of building blocks containing the leaf nodes of all 10,000 routes. Given the specifics of the PaRoutes dataset, the search algorithm has a maximum route length of 10 as this is the longest extracted route length from patents.

\subsection{Selected Approaches.} 

\begin{table}[h]
        \centering
        \caption{Selected single-step retrosynthesis models and multi-step synthesis planning algorithm}
       \label{tab:selected_approaches}
        \begin{tabular}{@{}m{8em} m{10em} m{19em}@{}}
        \toprule
\textbf{Task} & \textbf{Approach} & \textbf{Description} \\ \midrule
\multirow{2}{*}{\begin{tabular}[r]{@{}l@{}}Single-Step \\ Retrosynthesis\end{tabular}} & LocalRetro \cite{chenDeepRetrosyntheticReaction2021} &
Graph Neural Network predicting the application of local bond and atom templates \\ \cline{2-3}
& Chemformer \cite{irwinChemformerPretrainedTransformer2022} & Template-free sequence-to-sequence Transformer \\ \cline{2-3} 
& MHNreact \cite{seidlImprovingFewZeroShot2022} & Template-based information retrieval method relating products and template embeddings \\ \cline{2-3}
& AZF (Baseline) \cite{seglerNeuralSymbolicMachineLearning2017,genhedenAiZynthFinderFastRobust2020} & Default template-based method \\ \hline
\begin{tabular}[r]{@{}l@{}}Multi-Step \\ Synthesis Planning\end{tabular} & Retro* \cite{chenRetroLearningRetrosynthetic2020} & Best-first tree search algorithm leveraging A*-like pathfinding guided by the single-step model \\
        \bottomrule
       \end{tabular}
\end{table}

\textbf{Single-step Retrosynthesis.} We select three state-of-the-art single-step methods to evaluate within multi-step synthesis planning (Table \ref{tab:selected_approaches}). The selection is based on their top-n accuracy on the commonly used benchmarking dataset, USPTO-50k, ensuring to select models which employ the main research directions within the field, i.e., graph-based neural networks, sequence-to-sequence, and information retrieval. Where possible, we maintain the original implementation of the methods and only report deviations from this. 
 
LocalRetro \cite{chenDeepRetrosyntheticReaction2021} is a template-based method that uses local atom and bond templates. It applies a graph neural network to create embeddings for both atoms and bonds of a product, which are used in a classification task to predict appropriate templates and reaction centers jointly. Contrary to the original implementation of the method, for AZ-1M and AZ-18M we filter for a minimum template frequency of three to avoid an infeasible number of local atom and bond templates. 
 
Chemformer \cite{irwinChemformerPretrainedTransformer2022} is a template-free method based on a Transformer architecture that uses BART \cite{lewisBARTDenoisingSequencetoSequence2020} pre-training on molecular SMILES and is then fine-tuned on the retrosynthesis task. It uses product SMILES as input to predict reactant SMILES using beam-search. We set the beam size to 50.
 
MHNreact \cite{seidlImprovingFewZeroShot2022}, a template-based information retrieval approach, trains separate product and template encoders and uses modern Hopfield Networks \cite{ramsauerHopfieldNetworksAll2021} to relate products and template embeddings to find the most applicable reaction template. The original implementation uses all template embeddings simultaneously. However, due to large RAM requirements (>300GB) of this approach for USPTO-PaRoutes-1M, AZ-1M and AZ-18M, the templates are used in batches to train the model. Moreover, we apply a cut-off of a minimum of three template occurrences for AZ-1M and do not show results for AZ-18M as due to increased reaction diversity leading to a much larger number of templates requiring an unfeasible amount of memory.
 
Additionally, we include a simple template-based model as a baseline referred to as AZF, adapted from NeuralSym \cite{seglerNeuralSymbolicMachineLearning2017}, which is the default model in the most used public route planning software implementation AiZynthFinder \cite{genhedenAiZynthFinderFastRobust2020}. Noteworthy, this model architecture is also commonly used to benchmark novel multi-step search algorithms. Templates are extracted using the standard implementation of RDChiral \cite{coleyRDChiralRDKitWrapper2019} with a radius of two. Only templates with at least three occurrences are kept for USPTO-50k, USPTO-PaRoutes-1M, and AZ-1M, for AZ-18M templates with at least ten occurrences were kept, following \cite{genhedenAiZynthTrainRobustReproducible2023}.

\textbf{Multi-step Synthesis Planning.} For multi-step synthesis planning, we select Retro* \cite{chenRetroLearningRetrosynthetic2020} as the search algorithm used in all experiments. Retro* is a best-first tree search algorithm leveraging A*-like pathfinding guided by a neural network, where each algorithm iteration applies a single model call. We select Retro* as the multi-step algorithm since prior work shows minimal differences across multi-step algorithms \cite{trippReEvaluatingChemicalSynthesis2022}, though this is only shown for the common NeuralSym model architecture. Moreover, Retro* performs better than MCTS with state-of-the-art single-step retrosynthesis models, which require longer inference times \cite{hassenMindRetrosynthesisGap2022}. This performance difference is likely because Retro* does not require online planning for search tree traversal, limiting the number of single-step model calls required. Noteworthy, we defer from using a self-play dependent route planning algorithm, even though they have the highest reported benchmark performance \cite{liuRetrosyntheticPlanningDual2023} since self-play algorithms are not training data and single-step model agnostic, i.e., changes in stock or single-step model change the learned self-play tree traversal policy. This aspect is especially problematic for this work as every single-step model and data combination would require self-play training such that it would become unclear whether the single-step model or the self-play aspect is important for route planning. 
Furthermore, we use Retro* with no cost function, such that the reactant probability of the single-step model is the guiding probability in the tree search. The search goal of Retro* is to find synthesis routes that end in building block molecules, however, that information is not used to shape the reward, as in MCTS \cite{seglerPlanningChemicalSyntheses2018, genhedenAiZynthFinderFastRobust2020}, where the percentage of building block leaves is used to guide the tree search. Instead, the sole guidance of the tree search comes from the single-step model to prioritize reactions to explore. We defer from using the oracle function because it has shown little impact \cite{trippReEvaluatingChemicalSynthesis2022} and is trained on USPTO data, which could cause information leakage. For all searches, we use a maximum search time of 8 hours (28800 seconds) and 200 algorithm iterations. Furthermore, the top 50 reactions from the single-step model are added to the search tree at every iteration, deferring from using a cumulative probability cut-off. Moreover, unless otherwise stated, we use a maximum synthesis route length of 7 and the Zinc \cite{genhedenAiZynthFinderFastRobust2020} building block set consisting of 17,422,831 molecules.

\subsection{Implementation.} 
All single-step retrosynthesis models are incorporated into the AiZynthFinder \cite{genhedenAiZynthFinderFastRobust2020} synthesis planning framework using a newly developed common single-step model interface, ModelZoo. We extend AiZynthFinder such that any single-step model can be tested and used interchangeably within all implemented multi-step search algorithms. Where possible, the original single-step model code is used. All code will be made available on GitHub upon publication.

\subsection{Computational requirements.}
All single-step models for this work are trained on GPUs (Tesla V100). However, route planning is conducted on CPUs, given that insufficient GPUs are available for embarrassingly parallel evaluation of 10,000 molecules for each single-step model. In total, more than 1.5 million CPU hours were used to create the reported results.

\section{Results}
\subsection{Single-step retrosynthesis prediction}

\begin{figure}[hbt]
    \normalsize
    \centering
    \includegraphics[width=0.7\textwidth]{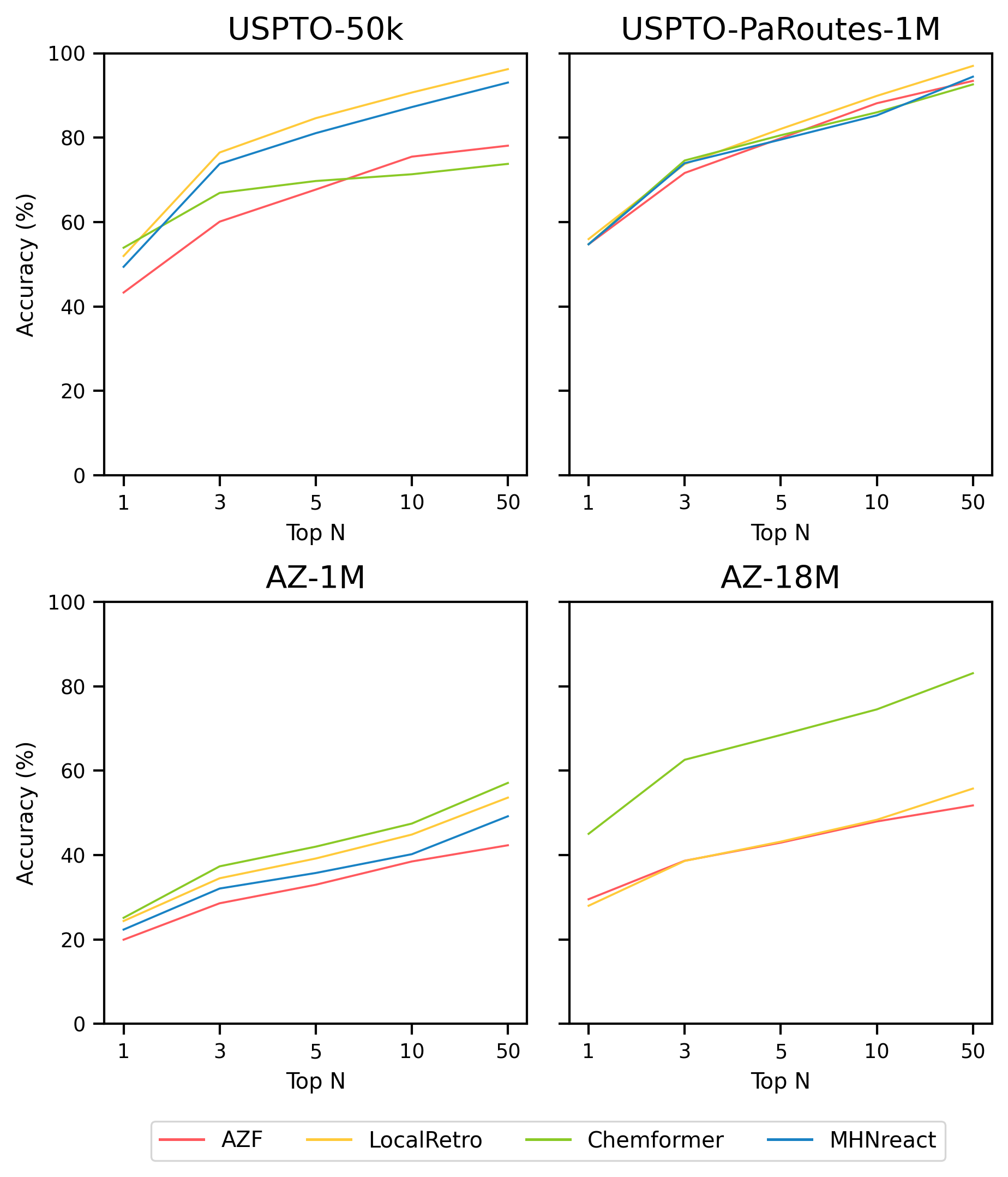} 
    \caption{Single-step Retrosynthesis Prediction Performance in terms of top-n accuracy for AZF, LocalRetro, Chemformer, and MHNreact on different datasets (USPTO-50k, USPTO-PaRoutes-1M, AZ-1M, AZ-18M) (see Supplementary Table \ref{supptab:single_step_performance}).}
    \label{fig:ssm_accuracy}
\end{figure}

\textbf{USPTO-50k.} As in the respective single-step retrosynthesis publications, the results on the USPTO-50k dataset, commonly used to benchmark and develop new single-step models \cite{zhongRecentAdvancesArtificial2023}, are reproducible. The best-performing methods are the state-of-the-art template-based methods (LocalRetro, MHNreact), which approach over 93\% accuracy by top-50 (Figure \ref{fig:ssm_accuracy}). Among those methods, LocalRetro is the best performing, closely followed by MHNreact. Chemformer, a template-free method, has the highest top-1 accuracy but stagnates as its performance does not increase with rising top-n. AZF is the worst-performing model until the top-10, where it outperforms Chemformer. However, AZF and Chemformer only reach a maximum of 77\% by top-50, an almost 19\% performance drop-off compared to LocalRetro and MHNreact.

\textbf{USPTO-PaRoutes-1M.} All models perform practically identically on the USPTO-PaRoutes-1M single-step dataset, with a maximum difference of ±4.6\% accuracy across all top-n (Figure \ref{fig:ssm_accuracy}), despite each approach employing different model architectures. At top-1, most models perform similarly, with LocalRetro outperforming the other models by 1\%. Within the top-3 accuracy, all state-of-art models (LocalRetro, Chemformer, MHNreact) maintain similar performance, whereas AZF performs slightly worse. By top-50, some slight differences are present, where LocalRetro is the best performing model, followed by MHNreact and the slightly worse performing AZF and Chemformer.

\textbf{AZ-1M.} In contrast to the comparably sized USPTO-PaRoutes-1M dataset, for AZ-1M the overall performance drops across all models (Figure \ref{fig:ssm_accuracy}). All three state-of-the-art models (LocalRetro, Chemformer, MHNreact) outperform AZF on all top-n accuracy levels. Both state-of-the-art template-based models perform similarly, where LocalRetro surpasses MHNreact as top-n increases. The template-free model, Chemformer, is the best-performing model throughout, though the difference is initially minimal, it becomes more pronounced across larger top-n. At top-50, Chemformer continues as the best-performing model, however it is closely followed by LocalRetro across all top-n.

\textbf{AZ-18M.} On the AZ-18M dataset, with an 18x increase of data compared to AZ-1M, Chemformer clearly outperforms the other models (Figure \ref{fig:ssm_accuracy}). At top-1, Chemformer already reaches an accuracy of 45.0\%, improving upon the other models by at least a +15.5\% margin. At top-50, Chemformer reaches 83.1\%, outperforming the next best model (LocalRetro) by +27.3\%. Noteworthy, both template-based methods (LocalRetro, AZF) perform similarly until top-10. Importantly, it was not possible to obtain results for MHNreact on AZ-18M due to the memory requirements of the method.

\subsection{Multi-step synthesis planning}
\subsubsection{Caspyrus10k}

\begin{table*}[hbt]
\small
\centering
\caption{\label{tab:caspyrus_ms} Multi-step synthesis planning performance on Caspyrus10k for different single-step models when trained on a diverse set of datasets. Measured by the success rate, indicating the number of molecules where a full synthesis route is found, the average number of solved routes, indicating the ability to produce synthesis route candidates, search times in seconds, and the average number of single-step model calls (see Supplementary Figure \ref{suppfig:ms_metrics} for distributions).
}
\begin{tabular}{@{}rrrcrrr@{}}\toprule
& & \multicolumn{1}{c}{Overall} & \phantom{}& \multicolumn{3}{c}{Average per Molecule} \\
\cmidrule{3-3} \cmidrule{5-7}
\textbf{Training Dataset} & \textbf{Model} & \textbf{Success Rate (\%)} && \textbf{Solved Routes} & \textbf{Search Time (s)} & \textbf{Model Calls} \\ \midrule
\multirow{4}{*}{\begin{tabular}[r]{@{}l@{}}USPTO-50k\end{tabular}}         & AZF        & 41.1             && 36.1          & 159             & 199                     \\
                 & LocalRetro & \textbf{74.1}             && 124           & 161             & 200                     \\
                 & Chemformer & 62.4             && 7.37          & 19051           & 177                     \\
                 & MHNreact   & 51.0             && 38.0          & 28958           & 99                    \\ \hline
\multirow{4}{*}{\begin{tabular}[r]{@{}l@{}}USPTO-PaRoutes-1M\end{tabular}}         & AZF        & 66.3             && 83.5          & 163             & 200                     \\
                 & LocalRetro & 86.0             && 324           & 1218            & 200                     \\
                 & Chemformer & \textbf{94.1}             && 463           & 28809           & 147                     \\
                 & MHNreact   & 64.6             && 215           & 28839           & 169                     \\ \hline
\multirow{4}{*}{\begin{tabular}[r]{@{}l@{}}AZ-1M\end{tabular}}             & AZF        & 73.5             && 124           & 168             & 200                     \\
                 & LocalRetro & 88.1             && 321           & 465             & 200                     \\
                 & Chemformer & \textbf{94.5}             && 358           & 29109           & 108                     \\
                 & MHNreact   & 56.0             && 77.0          & 29116           & 65                      \\ \hline
\multirow{3}{*}{\begin{tabular}[r]{@{}l@{}}AZ-18M\end{tabular}}            & AZF        & 76.2             && 154           & 154             & 199                     \\
                 & LocalRetro & 87.3             && 350           & 2736            & 200                     \\
                 & Chemformer & \textbf{90.9}             && 381           & 30212           & 75                     \\

\bottomrule
\end{tabular}
\end{table*}

Multi-step metrics of single-step models in synthesis planning are evaluated on Caspyrus10k, specifically route-finding success rate, average number of solved routes per molecule, average number of single-step model calls per molecule, and the average search time per molecule (see Methods). This establishes an overview of the capabilities of different models, trained on different datasets, across a large synthesizable chemical space.

\textbf{USPTO-50k.} For models trained on the USPTO-50k dataset, LocalRetro is the best-performing model with the highest success rate and average number of solved routes. Regarding success rate, a large disparity of ±32.0\% between the best-performing and worst-performing models is present. LocalRetro performs best, with a success rate of 74.1\%, followed by Chemformer, MHNreact, and AZF, with each model decreasing in performance by around 10\% from the previous one. The average number of solved routes per molecule also differs largely between the different single-step models, with the best-performing model producing almost 17x more solved routes than the worst-performing model. Again, LocalRetro performs best with 124 solved routes, followed by MHNreact, AZF, and Chemformer. In terms of single-step model calls, AZF, LocalRetro, and Chemformer approach the 200 model-call limit, yet there is a large disparity in search time. LocalRetro and AZF require only around 160 seconds per molecule, whereas Chemformer reaches an average search time of 5.3 hours (19,051 seconds). Lastly, despite reaching the search time limit, MHNreact has a considerably lower number of model calls. 

\textbf{USPTO-PaRoutes-1M.} Models trained on the USPTO-PaRoutes-1M dataset have considerable performance differences in synthesis planning, even though they perform similarly on the single-step test data (Figure \ref{fig:ssm_accuracy}). With the increased data volume, compared to USPTO-50k, all models solve a much larger portion of Caspyrus10k. The best-performing model in terms of success rate is Chemformer with 94.1\%, followed by LocalRetro, AZF, and finally MHNreact. Overall, the average number of solved routes is high for state-of-the-art single-step models. Chemformer finds, on average, 463 solved synthesis routes, followed by LocalRetro and MHNreact with 324 and 215, respectively. In comparison, the baseline AZF model finds only 83.5 solved routes per molecule. Concerning search time, Chemformer and MHNreact both exhaust the maximum search time, where neither reaches the maximum number of model calls. AZF is by far the fastest method, reaching 200 model calls in an average of 163 seconds. LocalRetro reaches the iteration limit within 1218 seconds on average, 7.5x slower than AZF but considerably faster than other state-of-the-art models.

\textbf{AZ-1M.} For AZ-1M, no clear performance improvement pattern is present in comparison to USPTO-PaRoutes-1M. In terms of success rate, AZF has a +7\% gain compared to USPTO-PaRoutes-1M, whereas Chemformer  and LocalRetro  maintain a very similar success rate. MHNreact, however, drops in route-finding success, reaching only 56.0\%. The average number of solved routes slightly increases for AZF compared to USPTO-PaRoutes-1M, whereas the performance decreases by 105 routes for Chemformer and more than halves for MHNreact. LocalRetro performs comparably with a minimal decrease of 3 solved routes. Regarding search time, both Chemformer and MHNreact exhaust the maximum search times, again not reaching the maximum number of single-step model calls. In fact, both models have a particularly low number of model calls, on average carrying out 108 model calls for Chemformer and 65 model calls for MHNreact. Both LocalRetro and AZF reach the maximum iteration limit, but LocalRetro is 2.77x slower.

\textbf{AZ-18M.} Finally, the success rate of models trained on the considerably larger AZ-18M dataset is comparable to the performance on AZ-1M with no changes beyond ± 3.6\%, even though the single-step performance can differ massively between both single-step datasets (Figure \ref{fig:ssm_accuracy}). Compared to AZ-1M, all models produce more solved routes. Chemformer solves the most routes per molecule, followed by LocalRetro and AZF. As for the search times, Chemformer once again reaches the time limit of 8 hours, whereas LocalRetro is considerably faster on average, beaten only by AZF. AZF and LocalRetro each reach the maximum iteration limit, whereas Chemformer only has 75 single-step model calls on average. Even though Chemformer success rate decreases, it can still produce the highest number of solved routes and the best success rate on AZ-18M.

\subsubsection{PaRoutes}

\begin{table*}[hbt]
\small
\centering
\caption{\label{tab:paroutes_ms} Multi-step Synthesis Planning performance on PaRoutes for different single-step models when trained on USPTO-PaRoutes-1M. Measured by the success rate, indicating the number of molecules where a full synthesis route is found, the average number of solved routes, indicating the ability to produce synthesis route candidates, search times in seconds, and the average number of single-step model calls (see Supplementary Figure \ref{suppfig:ms_metrics_paroutes} for distributions).
}
\begin{tabular}{@{}rrrcrrr@{}}\toprule
& & \multicolumn{1}{c}{Overall} & \phantom{}& \multicolumn{3}{c}{Average per Molecule} \\
\cmidrule{3-3} \cmidrule{5-7}
\textbf{Training Dataset} & \textbf{Model} & \textbf{Success Rate (\%)} && \textbf{Solved Routes} & \textbf{Search Time (s)} & \textbf{Model Calls} \\ \midrule
\multirow{4}{*}{\begin{tabular}[r]{@{}l@{}}USPTO-PaRoutes-1M\end{tabular}}   & AZF        & 97.1 && 159 & 153   & 200 \\
         & LocalRetro & 98.9 && 161 & 1067  & 200 \\
         & Chemformer & \textbf{99.7} && 524 & 28538 & 157 \\
         & MHNreact   & 91.1 && 173 & 28802 & 156 \\

\bottomrule
\end{tabular}
\end{table*}

Instead of evaluating the general route-finding abilities of single-step retrosynthesis models, PaRoutes focuses on the ability to recover gold-standard routes given a set of molecules and their predefined target building blocks. 
In terms of multi-step metrics, using the same evaluation as for Caspyrus10k, all models achieve an extremely high success rate of at least 91\% (Table \ref{tab:paroutes_ms}). In particular, AZF, LocalRetro, and Chemformer find solutions for practically all PaRoutes compounds. The three template-based methods (AZF, MHNreact, LocalRetro) produce a similar number of solved routes per molecule ranging between 159 and 173, whereas Chemformer surpasses these with an average of 524 solved routes per molecule (Table \ref{tab:paroutes_ms}, Supplementary Figure \ref{suppfig:ms_metrics_paroutes}). As already seen with Caspyrus10k, Chemformer and MHNreact reach the maximum search time of 8 hours without maxing out the single-step model calls. LocalRetro and AZF perform considerably faster, with AZF taking just 153 seconds on average to reach the maximum of 200 iterations.

\begin{figure}[hbt]
    \normalsize
    \centering
    \includegraphics[width=0.9\textwidth]{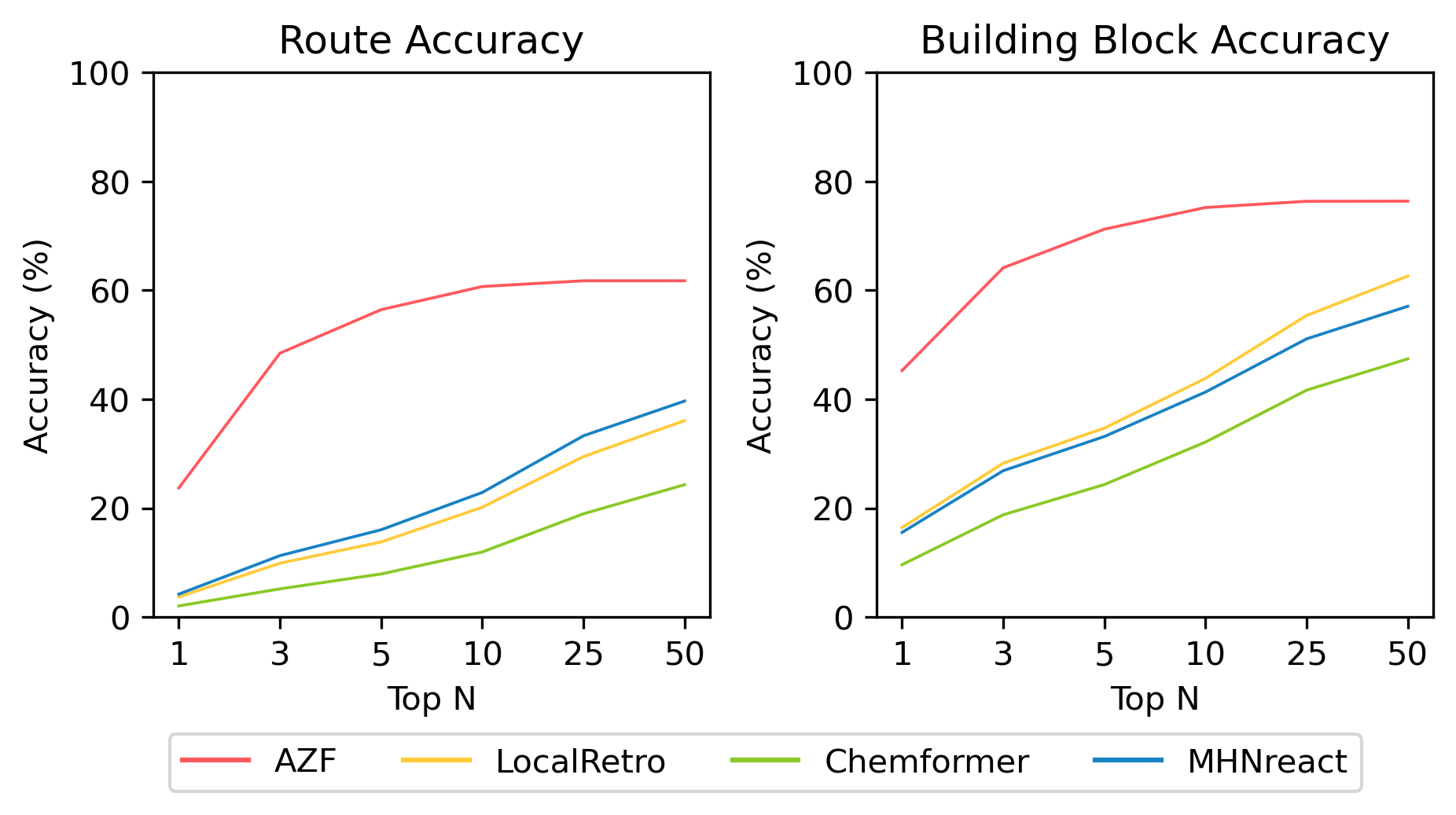} 
    \caption{Multi-step synthesis planning accuracy on PaRoutes gold-standard synthesis routes with different single-step models trained on USPTO-PaRoutes-1M. Route accuracy measures the ability to recover the correct synthesis route within top-n, whereas building block accuracy measures the ability to recover the correct building blocks while not considering reactions and intermediates (see Supplementary Table \ref{supptab:paroutes_performance}).}
    \label{fig:paroutes_accuracy}
\end{figure}

The route accuracy of the single-step model in synthesis planning measures how often the gold-standard synthesis route is recovered for a target molecule, where the selected n-1 set \cite{genhedenPaRoutesFrameworkBenchmarking2022} features only one retrosynthetic route per target-molecule. AZF has by far the best route accuracy overall, recovering 61.8\% of gold-standard routes within its top-50 predicted synthesis routes (23.7\% at top-1) (Figure \ref{fig:paroutes_accuracy}). Noteworthy, the performance plateaus after top-10 (at 60.7\%) and with little improvement at higher top-n. Both state-of-the-art template-based methods perform similarly across all top-n, but underperform compared to AZF by around -20\% (MHNreact: 39.7\%, LocalRetro: 36.1\%).The template-free Chemformer model is worst-performing across all top-n, reaching only 11.9\% by top-50. Noteworthy, the performance for all state-of-the-art models improves until the top-1000 (Supplementary Figure \ref{suppfig:ms_metrics_paroutes}), but never reaches the performance of AZF.

Considering the building block accuracy, which measures if the correct building blocks of the reference route are predicted while not considering the route reactions or intermediate molecules, considerable improvements for all models are present compared to the route accuracy. Within the top-50 synthesis route predictions, AZF correctly predicts the building blocks for 76.4\% of the gold-standard synthesis routes, a +14.6\% increase over its route accuracy. This improvement pattern is also present for the state-of-the-art models within the top-50 predicted synthesis routes, where all three state-of-the-art methods see a considerable improvement with at least a +17\% improvement between route and building block accuracy.

\section{Discussion}

Thus far, the task of retrosynthesis prediction is treated as two separate machine learning research fields. In this work, single-step retrosynthesis and multi-step synthesis planning are joined to analyze the impact of the single-step model on multi-step synthesis planning (Figure \ref{fig:graphic_abstract}). In particular, the focus is on vital aspects of synthesis planning, the single-step model, the multi-step search algorithm, and their domain-specific applicability.   

\subsection{Impact on single-step retrosynthesis prediction}
Considering the single-step retrosynthesis accuracies (Figure \ref{fig:ssm_accuracy}, Supplementary Table \ref{supptab:single_step_performance}), it can be stated that the default single-step retrosynthesis benchmark dataset, USPTO-50k, is problematic as there is no performance transfer of models between different datasets. A model performing well on the smaller 50k reaction dataset does not necessarily perform well on larger, more diverse datasets, as the ranking of the best-performing single-step model changes for every dataset. Generally, model performance increases, or stays comparable, with more data available. For instance, for USPTO-PaRoutes-1M, a superset of USPTO-50k with a larger number of reaction classes, the performance increases (AZF and Chemformer) or stays comparable (LocalRetro, MHNreact). This pattern is also present when comparing AZ-1M to its superset AZ-18M, where more data improves the performance slightly (LocalRetro) or substantially (Chemformer, AZF). For AZ-18M, the model with the highest jump in performance is the template-free Chemformer, reaching a top-50 accuracy of 83.1\% and substantially outperforming all other template-based methods by +27.3\%. Here it seems that the template-based nature of the other two models (AZF, LocalRetro) limits their ability to perform on the largest, most diverse dataset. This indicates that template-based methods may have reached a performance plateau due to not being to extrapolate beyond known templates, a limitation which is not present for the template-free Chemformer. Interestingly, for USPTO-50k, the template-free method is outperformed by all template-based methods at top-10 accuracy. Looking at the performance of AZF on AZ-18M, it is generally worse than shown in \cite{genhedenAiZynthTrainRobustReproducible2023}. 
The previous work uses a template-based stratified split for the hold-out split, leading to an even distribution of templates across the different splits and ensuring that every template is present in every split, which can benefit a template-based approach. However, in this work, we address the hold-out split by a strict random split on the reaction level, given the nature of the different single-step methods used. With increased data diversity, single-step performance diminishes for all models comparing the equally sized USPTO-PaRoutes-1M and AZ-1M (Figure \ref{fig:ssm_accuracy}). Data diversity is measured by the number of extracted unique reaction templates from the training splits of both datasets (USPTO-PaRoutes-1M: 314,959, AZ-1M: 439,618), representing different reaction ideas present in the respective datasets. This pattern is especially problematic, as USPTO-50k only includes ten reaction classes (USPTO-50k: 10,196 unique reaction templates).

Secondly, a novel benchmark is required for the single-step retrosynthesis research field, as methods developed for 50,000 data points are not easily transferable to real-world-sized datasets with millions of data points. Naturally, new methods should be developed using larger datasets that better encompass the size and diversity shown in real-world data since development for USPTO-50k limits their transferability (Figure \ref{fig:ssm_accuracy}). In terms of dataset size, all models require at least minor refactoring to run on larger datasets or do not scale beyond 1 million data points (MHNreact). Similarly, some USPTO-50k developed models do not conceptually consider the increase in reaction diversity in larger (real-world) datasets. For example, template-based models produce more templates with higher data diversity, requiring more template prediction classes in their classification tasks. Inherently, the number of classes a method can represent limits the number of different templates a method can predict. The solution to the diversity problem for those template-based methods is to remove templates occurring below a threshold and subsequently remove potential valid reaction predictions (see Methods). The natural exception are template-free methods as they are not constrained to reaction templates and show better scalability to more diverse data (Figure \ref{fig:ssm_accuracy}).
Noteworthy, USPTO-PaRoutes-1M \cite{genhedenAiZynthTrainRobustReproducible2023}, with its higher number of reactions and reaction diversity, is also not a perfect single-step model benchmark dataset since all single-step models perform comparably on it. Compared to the alternative public dataset USPTO-Full \cite{daiRetrosynthesisPredictionConditional2019}, the performance of all single-step models is much higher on USPTO-PaRoutes-1M, where LocalRetro has a more than +25\%   top-50 accuracy improvement \cite{zhongRecentAdvancesArtificial2023}.
The difference in single-step performance between USPTO-PaRoutes-1M and USPTO-Full and the equal performance on USPTO-PaRoutes-1M might be explainable by the underlying data sources and their respective preprocessing. USPTO-PaRoutes-1M is a superset of USPTO-Full, where the first contains USPTO grants and applications (3,748,191 total reactions) and the latter only USPTO grants (1,808,938 total reactions) \cite{thakkarDatasetsTheirInfluence2020}. In terms of preprocessing, USPTO-Full is noisier compared to USPTO-PaRoutes-1M as the latter applies extensive data cleaning and recreates and standardizes the atom-mapping between reactions with RXNMapper \cite{schwallerExtractionOrganicChemistry2021}. Naturally, given that all tested single-step models perform comparably on the most cleaned, standardized, publicly available dataset, the question remains whether a saturation point in single-step performance is reached on public data.

Directly inferring multi-step synthesis planning results from single-step retrosynthesis results is not possible since single-step model performance metrics do not directly transfer to multi-step route planning success. In fact, it is necessary to evaluate the performance of respective single-step models in a multi-step framework to evaluate their synthesis planning performance. In this study, single-step models performing equally well on the USPTO-PaRoutes-1M single-step task are performing vastly differently in multi-step synthesis planning. For example, Chemformer, compared to MHNreact,  has considerable differences in multi-step performance with a nearly ±30\% higher success rate and finding double the average number of solved routes per molecule (Table \ref{tab:caspyrus_ms}). Moreover, LocalRetro has a roughly +20\% higher success rate than AZF and finds 3.9x the number of solved routes. Looking at the disparities between USPTO-50k and other datasets, LocalRetro has the highest route-finding success of single-step models trained on the USPTO-50k dataset but is not the best-performing model when trained on larger datasets. Additionally, low single-step model performance on AZ-1M still leads to high multi-step performance. Here, the high diversity of reactions in AZ-1M, compared to the equally sized USPTO-PaRoutes-1M, might be the factor for the low single-step model performance. It seems that with fewer correctly predicted reactions, it is still possible to reach high multi-step performance. This aligns with prior works showing that most molecules can be addressed with relatively few reaction templates \cite{thakkarDatasetsTheirInfluence2020}.

\subsection{Impact on multi-step synthesis planning}
An important finding for multi-step synthesis planning is that the performance of route planning can be improved by merely switching out the single-step model, introducing novel reaction pathways to traverse the underlying reaction network (Table \ref{tab:caspyrus_ms}). In particular, huge success rate disparities are present within datasets, where the performance difference in finding a synthesis route between the best and worst models can be up as high as ±38.5\% (USPTO-50k: ±33.0\%, USPTO-PaRoutes-1M: ±29.5\%, AZ-1M: ±38.5\%, AZ-18: ±14.7\%).   This performance disparity pattern between the best and worst performing models trained on the same dataset is also present for the average number of solved routes per molecule, where the difference in solved routes ranges in the hundreds (USPTO-50k: ±117, USPTO-PaRoutes-1M: ±380, AZ-1M: ±281, AZ-18M: ±227). 
The availability of more reaction data can improve the success rate of route planning up to a certain level, where the largest jump is present between USPTO-50k and USPTO-PaRoutes-1M. Noteworthy, public data is on par with private data in terms of multi-step success rate for Chemformer and LocalRetro which have comparable performance when trained on USPTO-PaRoutes-1M or AZ-1M. However, for AZF, public datasets perform much worse as more reaction templates are extractable from private data \cite{genhedenAiZynthTrainRobustReproducible2023}. For MHNreact, private data even decreases the performance as the added complexity highly increases inference times, and only 65 single-step model calls are conducted in a generous 8-hour search window. 
The availability of more diverse reaction data can increase the average number of solved synthesis routes produced. Generally, we see that as reaction diversity of the single-step data increases so does the number of solved synthesis routes though eventually this performance stagnates or even worsens due to model architecture limitations. All models have either longer run times, if they reach the iteration limit, or a reduced number of single-step model calls, if they reach the time limit, reducing their potential to explore additional synthesis pathways. In the case of LocalRetro, where the minimum reaction template occurrence is increased from USPTO-PaRoutes-1M to AZ-1M from one to three due to an infeasible number of reaction template classes in the more diverse dataset, the search times massively decrease while even improving the success rate likely due to the decreased number of reaction templates. Finally, template-based models produce their respective most solved synthesis routes using the 18 million reaction dataset, AZ-18M. Chemformer, however, achieves less solved routes compared to USTPO-PaRoutes-1M as the number of single-step model calls is halved for the largest dataset, suggesting that the inference becomes slower with more diverse data.

Even though single-step retrosynthesis models improve the performance of route planning, they are generally not tailored to multi-step search algorithms. Single-step models have slow inference times that can deny high multi-step success rates, as few single-step model calls are possible within a set time limit and can also impede ad-hoc synthesis route generation. Attached to the inference problems of single-step models are the algorithmic properties of most multi-step algorithms. Though multi-step algorithms require single-step retrosynthesis models, they are generally developed to address a single molecule as a sequential next-disconnection prediction problem, with few exceptions \cite{xieRetroGraphRetrosyntheticPlanning2022}. Single-step models, however, are not optimized for this as they predict reactants for multiple different products simultaneously, typically in a joined GPU batch. Consequently, the combination of single-step and multi-step methods, though both thought for the task of retrosynthesis prediction, are currently not developed to be complementary to each other. Many of these models could massively improve their performance, particularly for the number of single-step calls and search time, by adapting the multi-step algorithm to suit the single-step model and vice-versa. Moreover, novel search algorithms, such as implementing asynchronous route planning, could have a substantial impact in this area.

\subsection{Impact on domain-specific applications}

Retrosynthesis prediction can be viewed as a domain-specific problem where the true objective of synthesis planning is to produce routes that can be used and tested experimentally. Given that there are multiple ways of synthesizing a molecule, the solution selected will often depend on the reaction preferences of the chemist and the desired route properties. As such, apart from the success rate and the number of solved routes, the route properties and their chemical validity are vital for the usefulness of the produced routes.

\begin{figure}[hbt]
    \normalsize
    \centering
    \includegraphics[width=\textwidth]{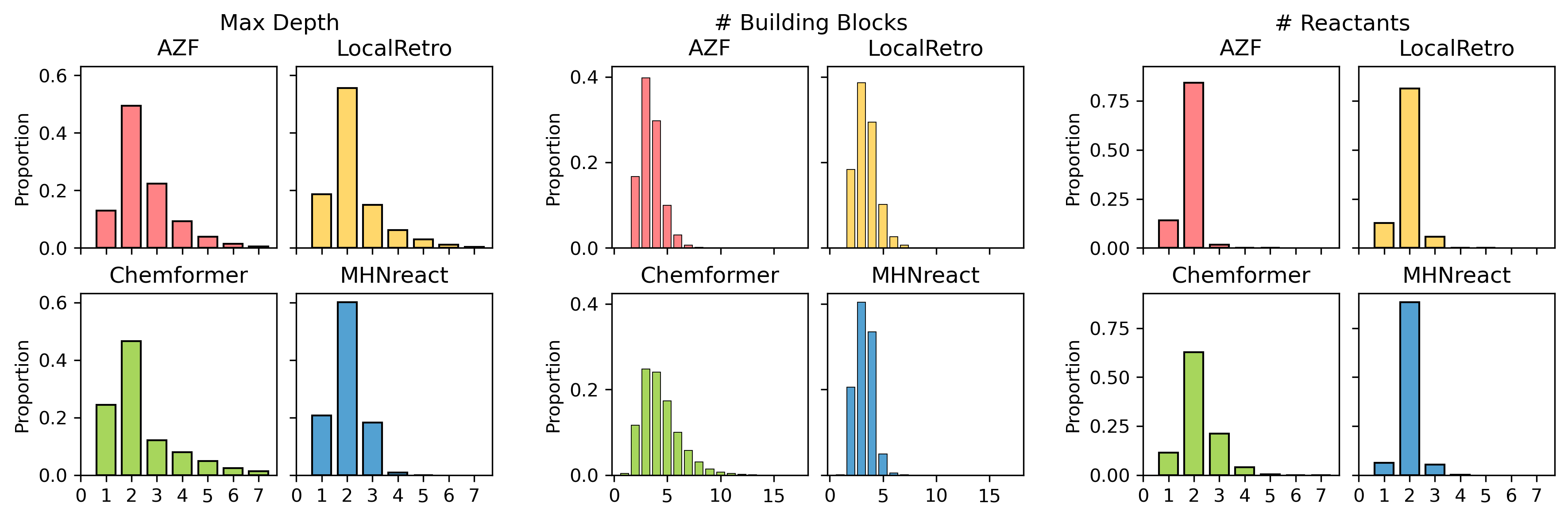} 
    \caption{Caspyrus10k route statistics of top-5 found synthesis routes by different single-step retrosynthesis models trained on USPTO-PaRoutes-1M. Shown are the maximum depth, referring to the longest linear path within the route, the number of building blocks within the route, and the number of reactants per route reaction.}
    \label{fig:route_characteristics}
\end{figure}

Generally, different models produce different route characteristics on Caspyrus10k (Figure \ref{fig:route_characteristics}), where the template-free method has noticeably different maximum route length, number of building blocks and number of reactants compared to the template-based methods. AZF and LocalRetro generally have very similar distributions across all characteristics, particularly in maximum route length where MHNreact has markedly shorter routes. Since MHNreact carries out a low number of single-step model calls within the maximum search time, it is likely that it is only able to address and solve short routes. Yet, Chemformer generally has a higher proportion of routes with a maximum depth of one, essentially directly predicting building blocks.  Additionally, Chemformer predicts a higher number of building blocks per route compared to all template-based methods, yet this effect is reduced with increased training data (Supplementary Figure \ref{suppfig:route_characteristics}).  Within the template-based methods we observe that the majority of reactions are bimolecular, producing two reactants, this is particularly true for MHNreact. Chemformer on the other hand predicts reactions which at times lead to four or more reactants.

 \begin{figure}[hbt]
    \normalsize
    \centering
    \includegraphics[width=0.8\textwidth]{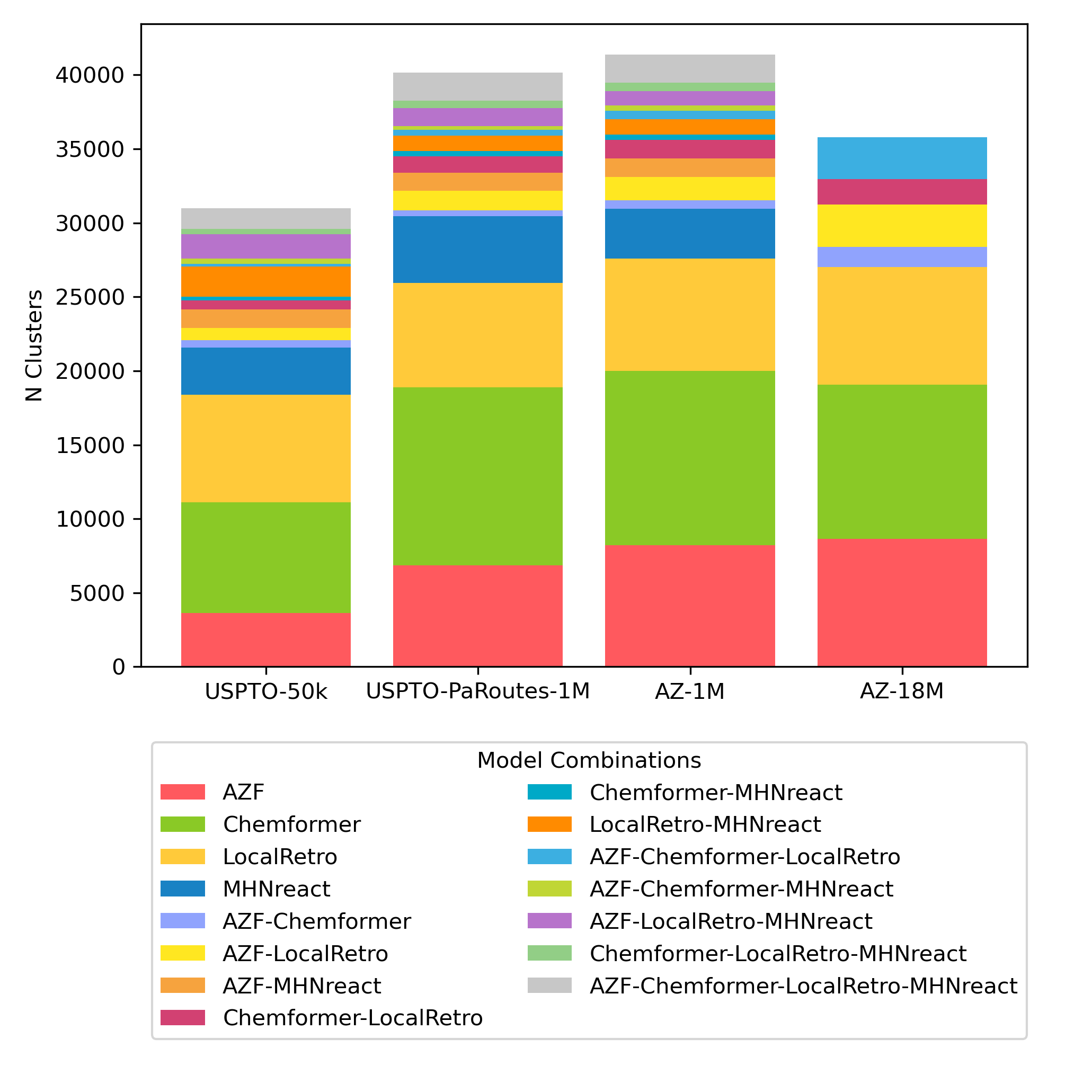} 
    \caption{Distribution and overlap of route clusters per single-step model and dataset when clustering with route-distance package \cite{genhedenClusteringSyntheticRoutes2021, genhedenFastPredictionDistances2022}. Clusters were calculated on a per molecule basis, N clusters shows the number of clusters which contained the stated combination of models.}
    \label{fig:route_clustering}
\end{figure}

Apart from looking at general route statistics of Caspyrus10k route planning results, we cluster the resulting synthesis routes to understand the relationship between different solved routes produced by distinct models within a reaction dataset. In detail, the approximated pairwise edit distance between solved synthesis routes of the top-5 predictions for each molecule is used to cluster with the route-distance package \cite{genhedenClusteringSyntheticRoutes2021, genhedenFastPredictionDistances2022}. Here, different single-step models produce unique route clusters when looking at the same training data, where routes produced by each model are generally unique to that model (Figure \ref{fig:route_clustering}). Noteworthy, routes produced by methods that rely on reaction templates (AZF, MHNreact, LocalRetro) tend to cluster together more frequently. Furthermore, models trained on AZ-18M tend to converge more regarding shared routes between models than models trained on USPTO-PaRoutes-1M. Nevertheless, the bulk of routes remains in unique clusters. Noteworthy, we check that the clustering patterns are also present when removing MHNreact (Figure \ref{suppfig:route_clustering_womhnreact}) to ensure that the missing MHNreact results for AZ-18M are not the sole reason for the difference between AZ-18M and the other datasets.

The availability of solved synthesis routes does not imply that those routes are also chemically valid. Validity can be assessed by comparing the produced routes of a single-step model to gold-standard routes as found in USPTO patents \cite{genhedenPaRoutesFrameworkBenchmarking2022} to indicate how valid the produced routes are. Generally, different single-step models are distinctive in their ability to reproduce gold-standard chemistry routes, i.e., route accuracy (Figure \ref{fig:paroutes_accuracy}). Surprisingly, there is no relationship between the multi-step success rate and the route accuracy of a single-step model. All models achieve at least 91\% success rates on PaRoutes target molecules (Table \ref{tab:paroutes_ms}) but differ considerably between route accuracies. AZF is the best-performing model regarding route accuracy, recovering 23.7\% of routes as the top-1 predicted synthesis route and 61.8\% within the top-50 predicted routes. In comparison, state-of-the-art models produce lower route accuracy, even if they produce high success rates. Within those state-of-the-art models, template-based models (LocalRetro and MHNreact) have a considerably higher route accuracy than the template-free approach Chemformer, yet still have a considerable gap in performance compared to the route accuracy of AZF.
 
Instead of predicting the correct gold-standard synthesis route, an easier task is to predict the right building blocks of the gold-standard route. This means that though the gold-standard route may not be entirely correctly predicted the building blocks are correctly predicted in the synthesis route, i.e., the order of the reactions may be incorrect or intermediate molecules are missing. For the easier task of predicting the correct building blocks, all models improve their performance compared to their respective route accuracy. However, the improvement between route accuracy and building block accuracy is much greater, compared to AZF, for state-of-the-art models that operate on local reaction templates (LocalRetro) or no templates at all (Chemformer), potentially meaning that they are more likely to skip vital aspects of the gold-standard synthesis routes in their route predictions rather than producing a distinct retrosynthesis route than the gold-standard route. Overall, the template-based AZF method performs best regarding building block accuracy.   
 
The performance difference on PaRoutes across different methods might be explainable by the allowed degree of chemical freedom of their respective model architectures. Template-based methods are more constrained by the reaction templates they apply, which are extracted from training reactions. With this constraint they are made to follow reaction pathways which are more chemically sound since their templates by definition, must be based on previous reactions. 
In comparison, the template-free Chemformer performs worst across both route and building block accuracy, potentially explained by the non-existent template guidance of the method allowing it to predict non-chemically sound reactions. Interestingly, this is in line with the divergence of Chemformer from general route statistics on Caspyrus10k, as the model predicts a much higher number of building blocks, multi-molecular reactions and routes that only consist of a singular reaction (Figure \ref{fig:route_characteristics}).

Generally, state-of-the-art approaches can provide a much larger set of route alternatives (Supplementary Figure \ref{suppfig:ms_metrics_paroutes}). This is also reflected in the PaRoutes route and building block accuracy, where AZF plateaus by top-10 accuracy, whereas state-of-the-art methods continue to increase their accuracy into very high top-n (Supplementary Figure \ref{suppfig:paroutes_accuracy_1k}). Given that state-of-the-art models produce more route alternatives, a future research direction, might be the best ranking of synthesis routes, as it can be assumed that desired routes are present within a large set of found synthesis routes.

 \begin{figure}[hbt]
    \centering
    \includegraphics[width=0.9\textwidth]{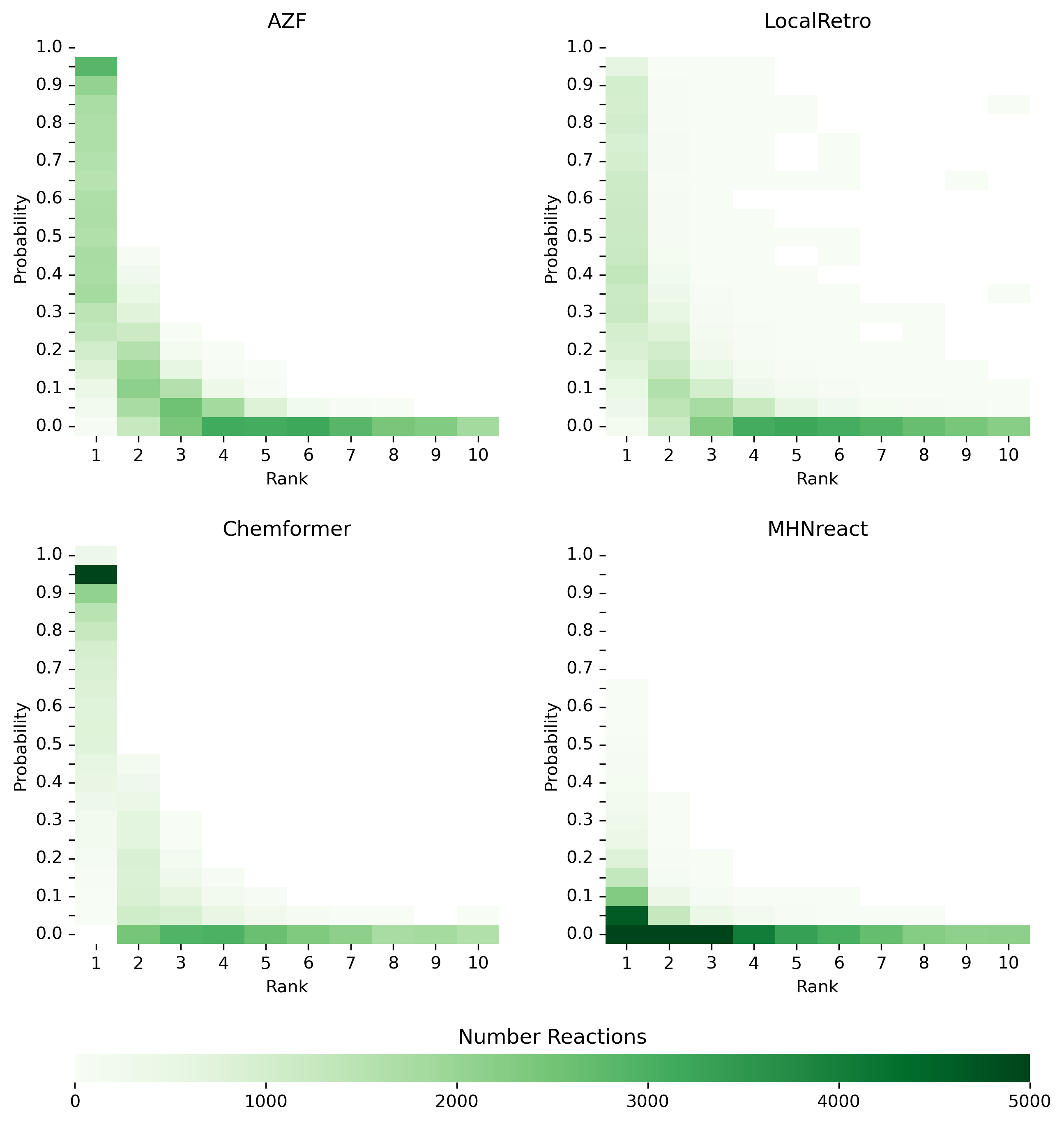} 
    \caption{Single-step model prior and rank distributions of reactions from the predicted and solved PaRoutes synthesis routes. A random sample of 100,000 reactions is extracted from the top-10 predicted routes (see Figure \ref{fig:paroutes_accuracy}) for each single-step retrosynthesis model trained on USPTO-PaRoutes-1M.}
    \label{fig:ssm_priors}
\end{figure}

An underlying assumption for single-step and multi-step synthesis planning is that the single-step model prior indicates the predicted chemical viability of a reaction for a molecule.  We assess this assumed relationship by extracting the predicted reaction probabilities and their respective rank for reactions from the top-10 solved routes of the PaRoutes benchmark dataset. We then select a random subset of 100,000 reactions for each model. Surprisingly, the single-step model prior distributions (Figure \ref{fig:ssm_priors}) show that for all models that there is no clear connection between the single-step reaction priors and the ability to find a solved synthesis routes as reactions of solved synthesis routes contain both low probability reactions and low prediction rank. Furthermore, models with a smoother progression between probabilities of higher and lower ranked disconnections (AZF, LocalRetro, MHNreact) tend to perform better at recovering gold-standard routes (Figure \ref{fig:paroutes_accuracy}). In contrast, a more skewed, overconfident, distribution towards top-1 predictions tends to perform worse (Chemformer). 

Though the routes found within the top-10 predicted routes use reactions with very low reaction probabilities, gold-standard routes are generally only found within the top-5 predicted reactions (Supplementary Figure \ref{suppfig:paroutes_accuracy_1k}). This suggests that routes with reactions ranked outside the top-5 predicted reactions, though leading to building blocks, produce non-viable route reactions (Figure \ref{fig:ssm_priors}). The presence of these low-probability reactions can be explained by the search algorithm ranking possible synthesis routes by their ability to reach building blocks and their overall route length. In the tree search itself, the search algorithm prioritizes short and solved routes, which might also include reactions with low probabilities as the overall search goal is to find a synthesis route ending in purchasable building blocks. The effect of low-probability reactions is enforced by adding 50 reactions to the search tree at every time step, even if those disconnections have low probabilities. 
Noteworthy, it is likely that the tree search algorithm explores those low-probability reactions when the high-probability disconnections are already explored. 
However, given the overall distribution of reaction priors (Figure \ref{fig:ssm_priors}), this approach might not be desired for future synthesis planning search algorithms. Furthermore, in future work, it could be interesting to analyze how the synthesis planning results differ when applying only the top-5 predicted reactions, consequently limiting the breadth of the search tree. Given that gold-standard routes are only found within the top-5 predicted routes (Supplementary Figure \ref{suppfig:ms_priors_accurate}), it opens the question if the resulting synthesis routes are closer to human-desired routes.
 
When discussing gold-standard synthesis routes, it is important to point out that a gold-standard route is only one way of synthesizing a desired molecule and other valid synthesis routes might also be possible. However, a good synthesis planning application should be able to prioritize real-world routes from a set of all potential routes, even if the favored chemical reactions change over time. Not finding the real-world routes entirely, yet identifying the correct building blocks, indicates that the produced synthesis routes are invalid or potentially missing vital parts of the synthesis route to be directly useful in an experimental setting. Naturally, there is a clear connection between the ability to recover gold standard routes and the ability to predict solved routes at all. High success rates produce route candidates that might be potential real-world synthesis routes but need to consider chemical validity. Because of this lack of validity, candidates are currently treated as initial retrosynthetic ideas. For a real improvement in the field of retrosynthesis, one of the essential questions, beyond improving the generation of possible solved route candidates, is how to evaluate and improve the chemical validity of generated synthesis routes. For this, it is vital to introduce reagents, conditions and yields into synthesis planning in the future and address the chemical feasibility of the generated routes. Though there is currently a lack of in-silico synthesis feasibility evaluation, as methods like round-trip accuracy \cite{schwallerPredictingRetrosyntheticPathways2020} only measure if the product is recoverable from the reactants and do not consider full chemical validity, given that retrosynthesis methods do not produce the relevant reagents and conditions required. Newer works have attempted to address this problem by predicting all required components \cite{kreutterMultistepRetrosynthesisCombining2023}. Chemical validity, however, could potentially be addressed with new advancements in the field, such as molecular dynamics or quantum chemistry prediction. 

Finally, when selecting the single-step retrosynthesis model for route planning, there are trade-offs between different desired search properties, as no approach outperforms all others if one uses a large enough dataset like USPTO-PaRoutes-1M. Clearly, there is a single-step performance advantage of template-free single-step models on large, heterogenous reaction data. However, this advantage comes at the cost of inference speed at multi-step synthesis planning, where template-based models are generally preferred as they can perform over 200-fold faster than template-free. If the overall goal of synthesis planning is a high success rate with a high average number of produced solved routes while accommodating long search times and a high divergence from reference routes, then the template-free approach, Chemformer, may be relevant. With a slightly lower success rate and average number of solved routes but much shorter runtimes and medium divergence from reference routes the successful state-of-the art template-based model, LocalRetro, is of interest. For very short run times and low divergence from reference routes yet lower success rate and an average number of solved routes, the default single-step retrosynthesis model, AZF, will be of use. Future developed models can aim to address a combination of these goals. 

One of the underlying problems in the field is that benchmarking different single-step retrosynthesis models within synthesis planning is time- and resource-intensive. However, to facilitate such benchmarking in the future, we analyze the variance of different subsample sizes of the Caspyrus10k multi-step synthesis dataset such that an approximation of the results can be carried out in lieu of running the full datasets for faster benchmarking/prototyping (see Supplementary Tables \ref{supptab:caspyrus_ms_subsample_100}- \ref{supptab:caspyrus_ms_subsample_5000}). In detail, we repeatedly randomly subsample a subset of molecules (100, 500, 1000, 5000 molecules) and measure the mean and standard deviation across 1000 subsamples (sampling without replacement). Given that the standard deviation is reasonably small for a sample size of 1000 molecules (see Supplementary Table \ref{supptab:caspyrus_ms_subsample_1000}), we provide a selected set of 1000 molecules if a full evaluation is not feasible (see Supplementary Table \ref{supptab:caspyrus_ms_selected_1000_mol_subsample}).

Noteworthy, this work only explores three state-of-the-art and a common baseline single-step retrosynthesis models, and even though representative of the common research directions, gives us only a snapshot of possible single-step and multi-step retrosynthesis combinations.

\section{Conclusion}
In this work, we create the first in-depth study combining state-of-the-art single-step retrosynthesis with multi-step synthesis planning, analyzing the gains and pitfalls of combining the two research fields. We find that there is generally no direct relationship between high single-step performance and successfully finding synthesis routes, both for publicly available and proprietary datasets, emphasizing the need to develop and evaluate single-step retrosynthesis models in a multi-step synthesis planning framework. Moreover, we show that the default single-step retrosynthesis benchmark dataset, USPTO-50k, is insufficient as methods developed for this small, homogenous dataset are not transferable to real-world, larger, and more diverse datasets. This is true for both single-step performance, where performance rankings between models are not transferable, and scalability, where model implementations are not transferable.

For multi-step synthesis planning, we show that the single-step model is an essential but thus far ignored aspect of the search algorithm. By merely changing the single-step retrosynthesis model it is possible to improve route-finding success by up to +28\%, reaching success rates above 90\% compared to the commonly used baseline model, when trained on the same reaction datasets. Furthermore, we show that every single-step model produces unique synthesis routes when used in multi-step synthesis planning, and each single-step model also differs in important aspects such as route-finding success, the average number of found synthesis routes, search times, and chemical validity. To summarize, we show that the combination of single-step retrosynthesis prediction and multi-step synthesis planning is a crucial aspect when developing future methods.

\section*{Acknowledgements}
This study was partially funded by the European Union’s Horizon 2020 research and innovation program under the Marie Skłodowska-Curie Innovative Training Network European Industrial Doctorate grant agreement No. 956832 “Advanced machine learning for Innovative Drug Discovery”. Parts of this work were performed using the ALICE compute resources provided by Leiden University.

\bibliographystyle{IEEEtran}
\bibliography{models_matter}

\newpage
\section*{Supporting Information}
\setcounter{table}{0}
\setcounter{figure}{0}
\renewcommand{\thetable}{S\arabic{table}}
\renewcommand{\thefigure}{S\arabic{figure}}
\renewcommand{\thesubsection}{\Alph{subsection}}

\subsection{Single-step retrosynthesis prediction}

\begin{table*}[hbt]
\small
\centering
\caption{\label{supptab:single_step_performance} Single-step Retrosynthesis Prediction Top-n Accuracy for AZF, LocalRetro, Chemformer, and MHNreact on the respective test sets of a dataset (USPTO-50k, USPTO-PaRoutes-1M, AZ-1M, AZ-18M).
}
\begin{tabular}{@{}rrrrrrrr@{}}\toprule
& & \multicolumn{5}{c}{Top-N Accuracy (\%)} \\
\cmidrule{3-7}
\textbf{Training Dataset} & \textbf{Model} & \textbf{Top-1} & \textbf{Top-3} & \textbf{Top-5} & \textbf{Top-10} & \textbf{Top-50}  \\ \midrule
\multirow{4}{*}{\begin{tabular}[r]{@{}l@{}}USPTO-50k\end{tabular}} & AZF & 41.6 & 62.5 & 69.5 & 75.8 & 77.4 \\
                 & LocalRetro & 52.0 & 76.5 & 84.6 & 90.7 & 96.2 \\
                 & Chemformer & 53.9 & 66.9 & 69.7 & 71.3 & 73.8 \\
                 & MHNreact   & 49.4 & 73.8 & 81.1 & 87.3 & 93.1 \\ \hline
\multirow{4}{*}{\begin{tabular}[r]{@{}l@{}}USPTO-PaRoutes-1M\end{tabular}} & AZF & 54.7 & 71.6 & 79.9 & 88.2 & 93.5 \\
                 & LocalRetro & 56.0 & 73.7 & 82.1 & 89.9 & 97.0 \\
                 & Chemformer & 54.8 & 74.6 & 80.6 & 86.0 & 92.6 \\
                 & MHNreact   & 54.7 & 74.0 & 79.5 & 85.3 & 94.5 \\ \hline
\multirow{4}{*}{\begin{tabular}[r]{@{}l@{}}AZ-1M\end{tabular}} & AZF & 19.9 & 28.6 & 33.0 & 38.5 & 42.3 \\
                 & LocalRetro & 24.4 & 34.5 & 39.2 & 44.9 & 53.6 \\
                 & Chemformer & 25.1 & 37.3 & 42.0 & 47.5 & 57.1 \\
                 & MHNreact   & 22.3 & 32.1 & 35.8 & 40.2 & 49.2 \\ \hline
\multirow{4}{*}{\begin{tabular}[r]{@{}l@{}}AZ-18M\end{tabular}} & AZF & 29.5 & 38.7 & 42.9 & 48.0 & 51.8 \\ 
                 & LocalRetro & 28.0 & 38.6 & 43.2 & 48.4 & 55.8 \\ 
                 & Chemformer & 45.0 & 62.6 & 68.5 & 74.5 & 83.1 \\ 
                 & MHNreact & - & - & - & - & - \\ 

\bottomrule
\end{tabular}

\end{table*}

\clearpage
\subsection{Multi-step synthesis planning}
\subsubsection{Caspyrus10k}
\begin{figure}[!hbt]
    \centering
    \begin{subfigure}[t]{0.3\textwidth}
         \centering
         \includegraphics[width=\textwidth]{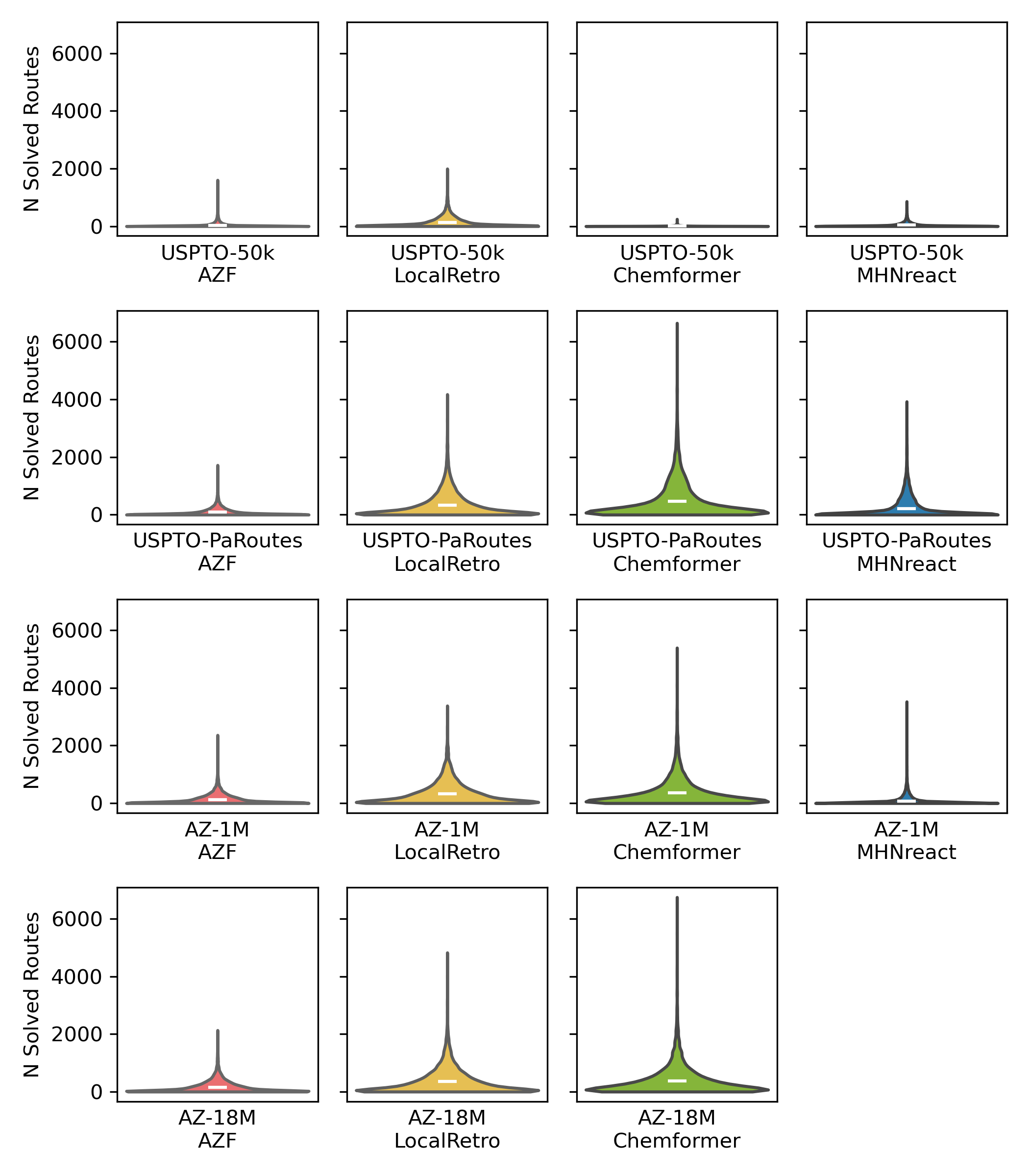}
         \caption{Solved routes}
         \label{suppfig:solved_routes}
     \end{subfigure}
     \hfill
     \begin{subfigure}[t]{0.3\textwidth}
         \centering
         \includegraphics[width=\textwidth]{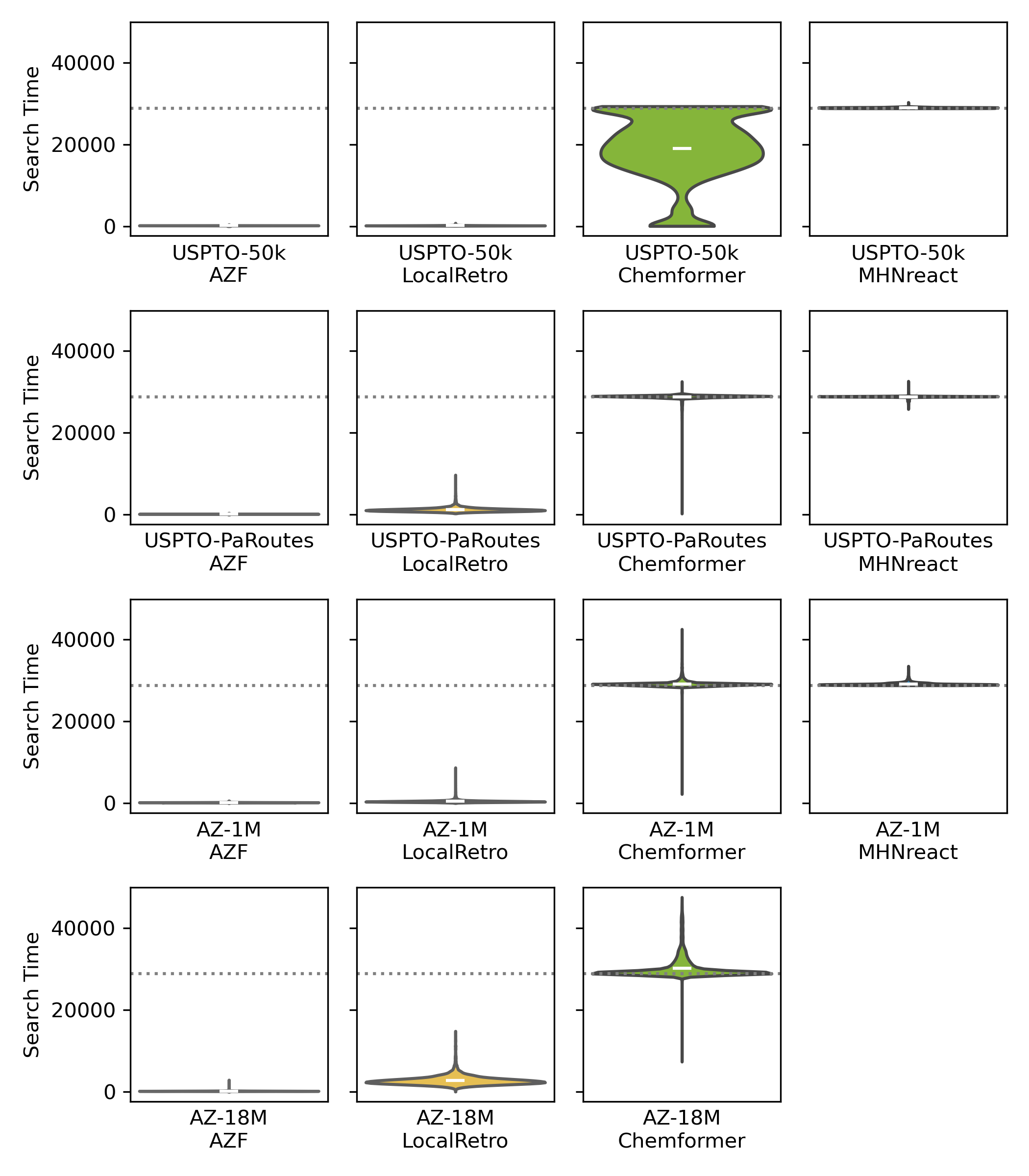}
         \caption{Search times}
         \label{suppfig:search_time}
     \end{subfigure}
     \hfill
     \begin{subfigure}[t]{0.3\textwidth}
         \centering
         \includegraphics[width=\textwidth]{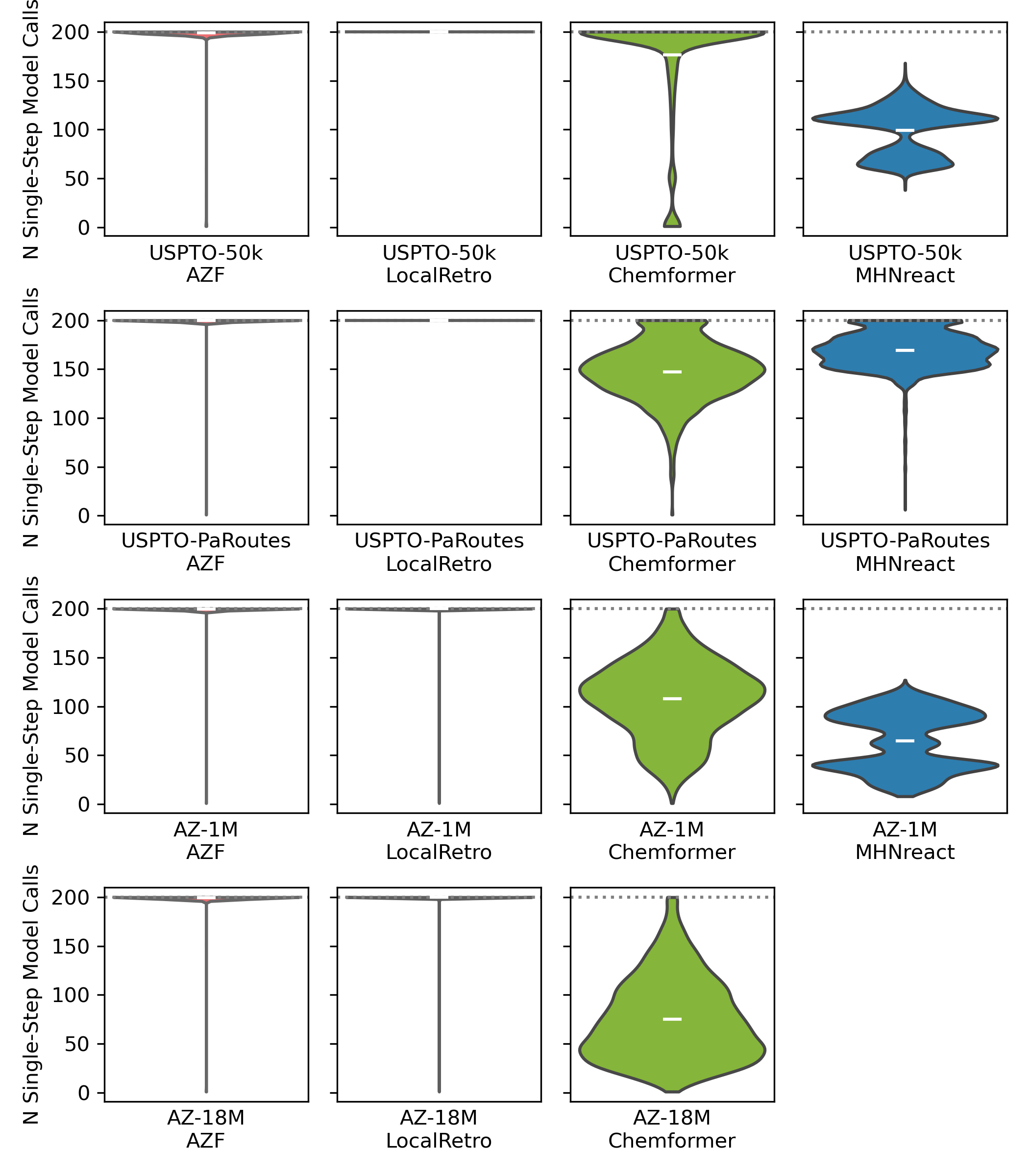}
         \caption{Single-step model calls}
         \label{suppfig:ssm_model_calls}
     \end{subfigure}
     \hfill
     
\caption{Distributions of solved routes (\subref{suppfig:solved_routes}), search time (\subref{suppfig:search_time}) and single-step model calls (\subref{suppfig:ssm_model_calls}) for synthesis planning results for all training datasets evaluated on Caspyrus10k. The dashed line indicates the respective limits set in algorithm search settings. The white line indicates the mean across all molecules for the shown model-training set combination.}
\label{suppfig:ms_metrics}

\end{figure}

\begin{figure}[hbt]
    \centering
    \begin{subfigure}[b]{0.3\textwidth}
         \centering
         \includegraphics[width=\textwidth]{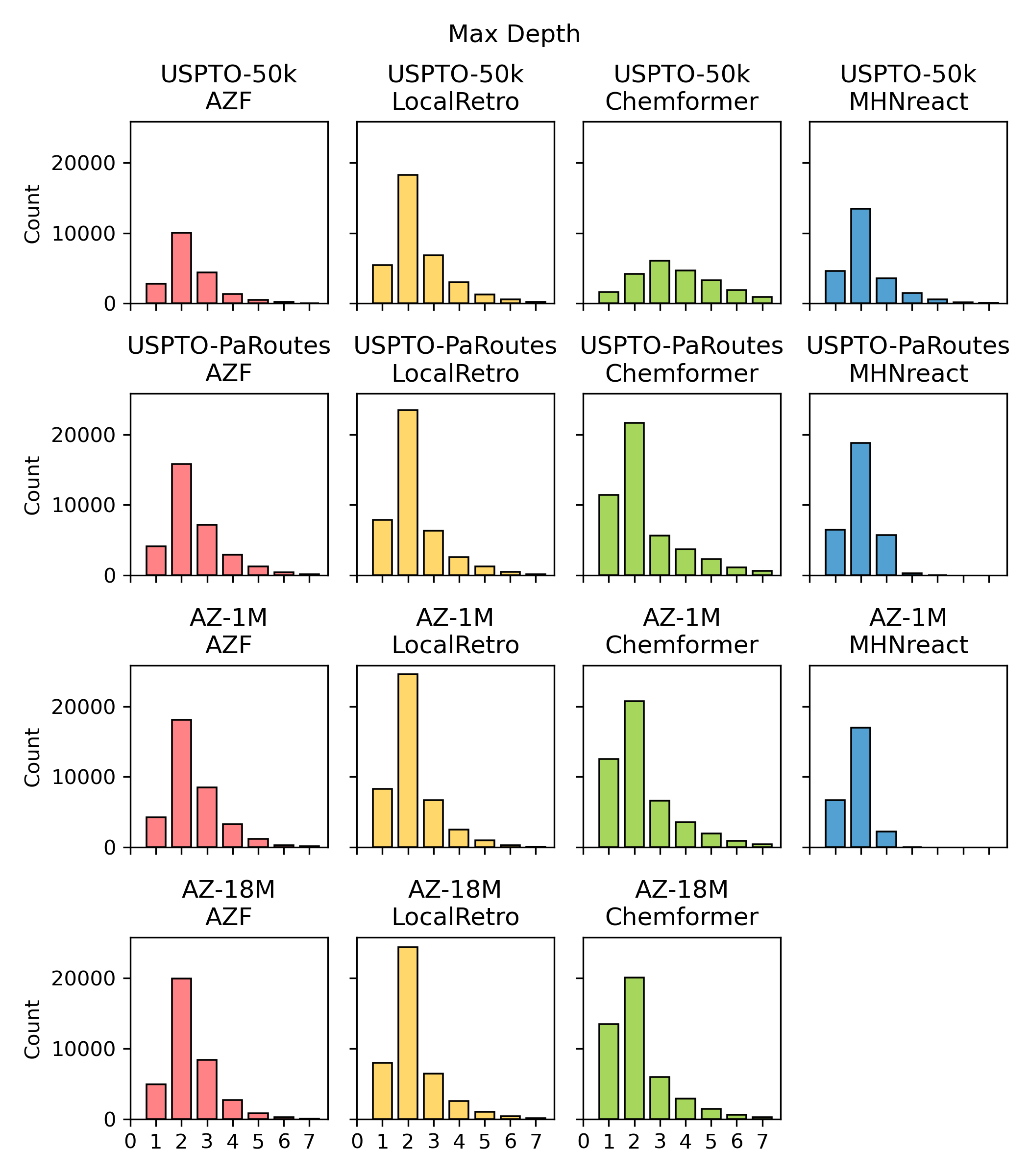}
         \caption{Maximum depth of routes}
         \label{suppfig:paroutes_route_depth}
     \end{subfigure}
     \hfill
     \begin{subfigure}[b]{0.3\textwidth}
         \centering
         \includegraphics[width=\textwidth]{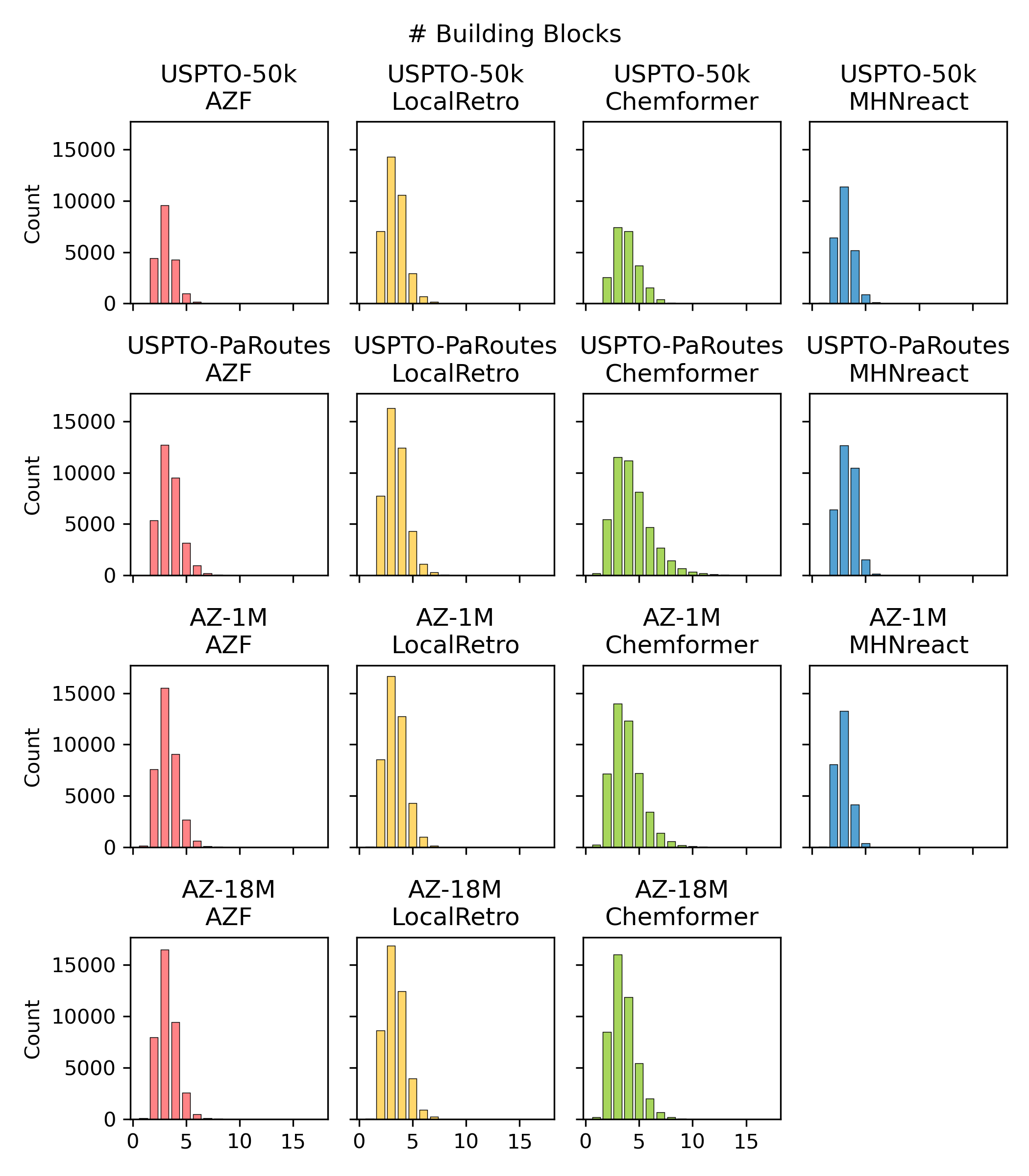}
         \caption{Building blocks per route}
         \label{suppfig:paroutes_building_blocks}
     \end{subfigure}
     \hfill
     \begin{subfigure}[b]{0.3\textwidth}
         \centering
         \includegraphics[width=\textwidth]{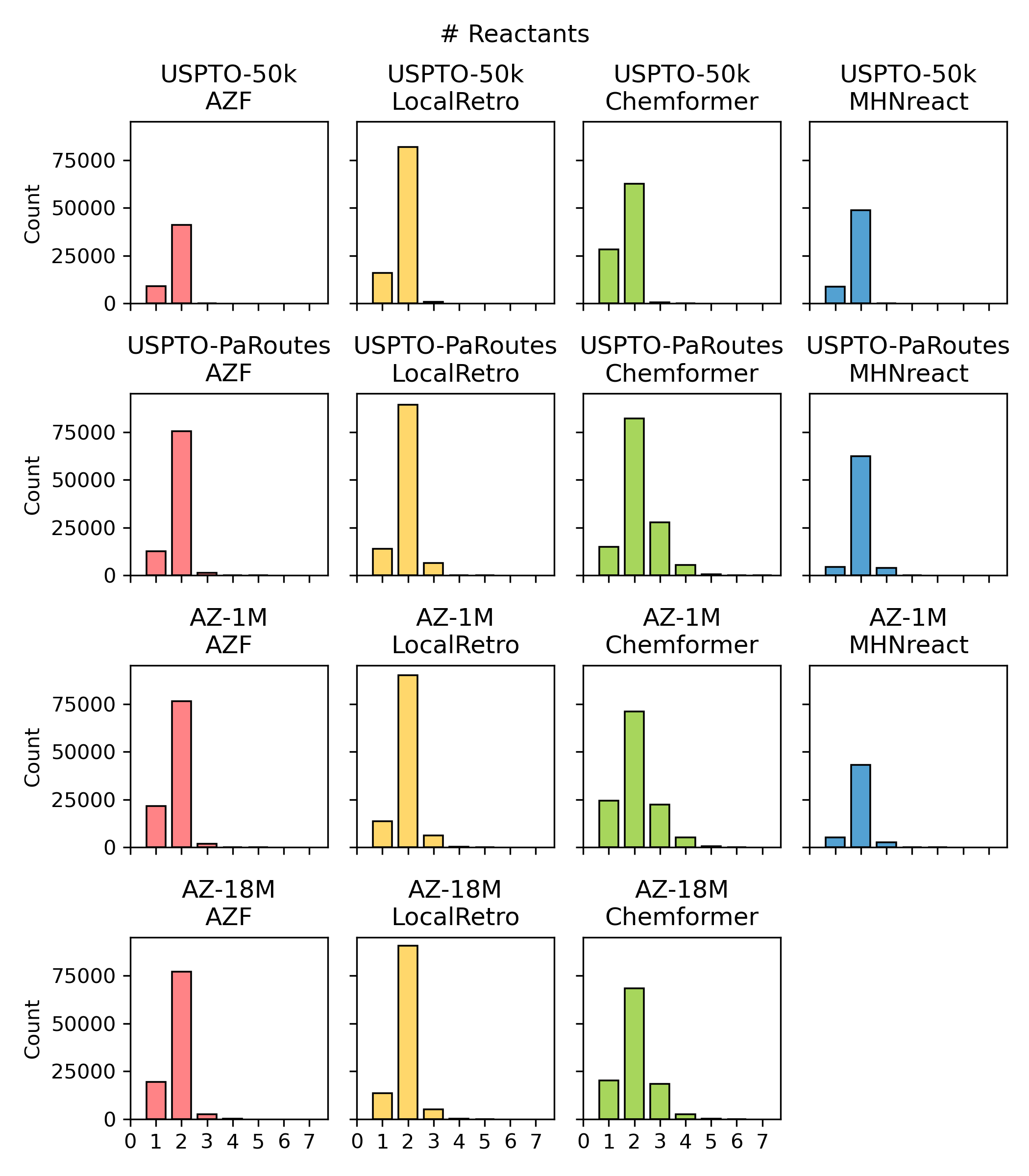}
         \caption{Number of reactants per reaction}
         \label{suppfig:paroutes_reactants}
     \end{subfigure}
     \hfill
     
    \caption{Statistics of top-5 found synthesis routes  on Caspyrus10k by different single-step retrosynthesis models for all datasets. Shown are the maximum depth (\subref{suppfig:paroutes_route_depth}), referring to the longest linear path within the route, the number of building blocks within the route (\subref{suppfig:paroutes_building_blocks}), and the number of reactants per route reaction (\subref{suppfig:paroutes_reactants})}
    \label{suppfig:route_characteristics}
\end{figure}

\begin{figure}[h]
    \centering
    \includegraphics[width=0.9\textwidth]{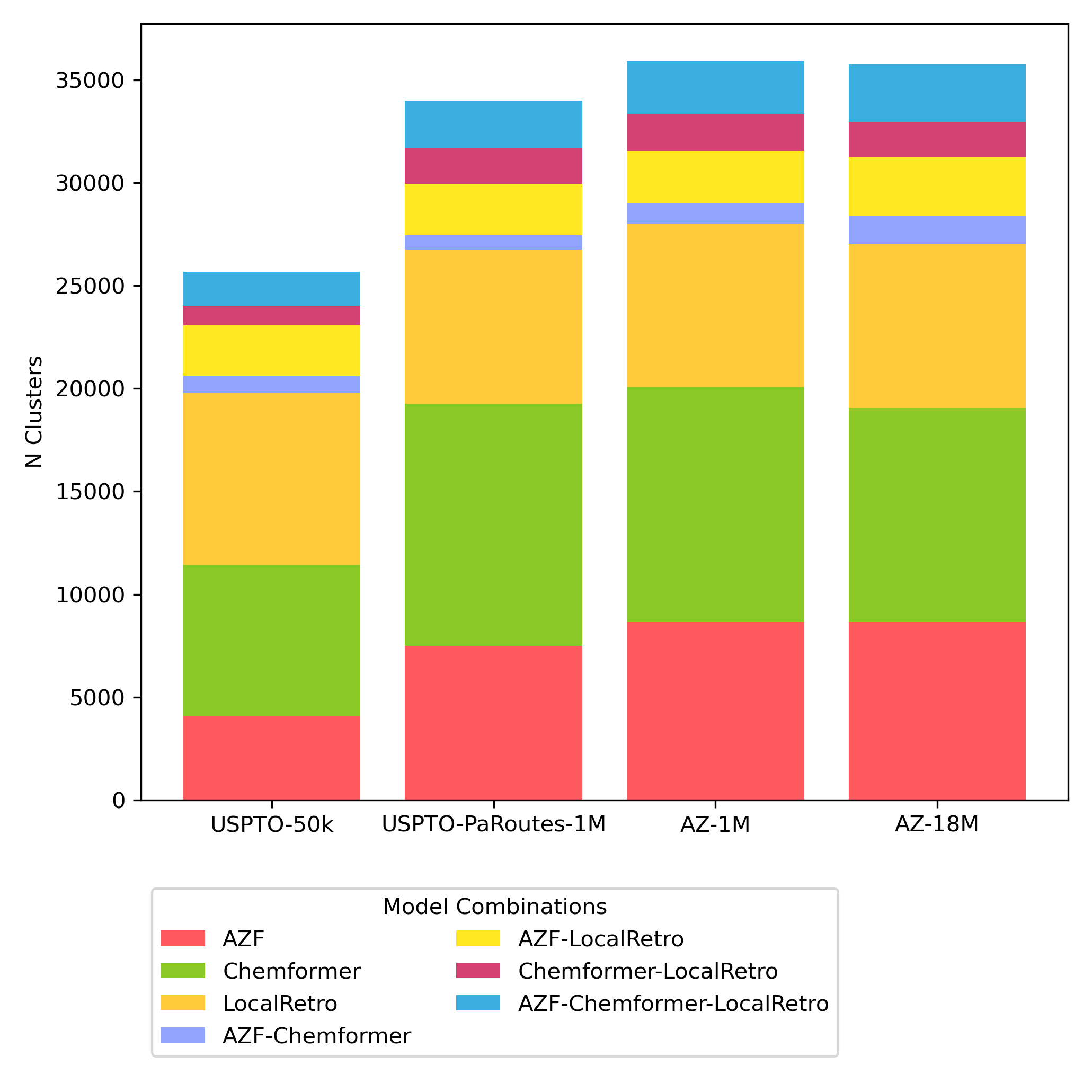} 
    \caption{Distribution and overlap of route clusters per single-step model (excluding MHNreact) and dataset when clustering with route-distance package \cite{genhedenClusteringSyntheticRoutes2021, genhedenFastPredictionDistances2022}. Clusters were calculated on a per molecule basis, N clusters shows the number of clusters which contained the stated combination of models.}
    \label{suppfig:route_clustering_womhnreact}
\end{figure}

\clearpage
\subsubsection{Caspyrus10k Subsampling}

\begin{table*}[hbt]
\small
\centering
\caption{\label{supptab:caspyrus_ms_subsample_100} Multi-step synthesis planning metrics for a subsample size of 100
Caspyrus10k molecules. The performance is measured for each single-step model and dataset by randomly subsampling 1000 times with the subsample size (sampling without replacement). For each subsample, the same molecules are used across single-step models and datasets.
}
\begin{tabular}{@{}rrrcrrr@{}}\toprule
& & \multicolumn{1}{c}{Overall} & \phantom{}& \multicolumn{3}{c}{Average per Molecule} \\
\cmidrule{3-3} \cmidrule{5-7}
\textbf{Training Dataset} & \textbf{Model} & \textbf{Success Rate (\%)} && \textbf{Solved Routes} & \textbf{Search Time (s)} & \textbf{Model Calls} \\ \midrule
\multirow{4}{*}{\begin{tabular}[r]{@{}l@{}}USPTO-50k\end{tabular}}         
                 & AZF        & 41.2 ± 5.0 && 36.6 ± 8.6 & 159 ± 2 & 198 ± 1\\
                 & LocalRetro & 74.2 ± 4.3 && 124 ± 18 & 160 ± 5 & 200 ± 0\\
                 & Chemformer & 62.5 ± 4.7 && 7.28 ± 1.51 & 19043 ± 789 & 176 ± 5 \\
                 & MHNreact   & 51.0 ± 5.0 && 38.8 ± 7.9 & 28956 ± 10 & 99.4 ± 2.4 \\ \hline
\multirow{4}{*}{\begin{tabular}[r]{@{}l@{}}USPTO-PaRoutes-1M\end{tabular}}         
                 & AZF        & 66.6 ± 4.8 && 84.2 ± 13.1 & 162 ± 1 & 199 ± 0\\
                 & LocalRetro & 86.3 ± 3.3 && 326 ± 42 & 1217 ± 49 & 200 ± 0\\
                 & Chemformer & 94.2 ± 2.4 && 464 ± 60 & 28811 ± 95 & 147 ± 2\\
                 & MHNreact   & 64.9 ± 4.7 && 215 ± 36 & 28839 ± 24 & 169 ± 1\\ \hline
\multirow{4}{*}{\begin{tabular}[r]{@{}l@{}}AZ-1M\end{tabular}}             
                 & AZF        & 73.7 ± 4.4 && 124 ± 17 & 168 ± 4 & 199 ± 0\\
                 & LocalRetro & 88.2 ± 3.1 && 322 ± 38 & 464 ± 34 & 199 ± 0\\
                 & Chemformer & 94.6 ± 2.3 && 360 ± 44 & 29110 ± 68 & 107 ± 3\\
                 & MHNreact   & 56.0 ± 5.1 && 77.2 ± 16.9 & 29114 ± 33 & 64.6 ± 3.0\\ \hline
\multirow{3}{*}{\begin{tabular}[r]{@{}l@{}}AZ-18M\end{tabular}}            
                 & AZF        & 76.4 ± 4.1 && 154 ± 21 & 153 ± 4 & 199 ± 1\\
                 & LocalRetro & 87.4 ± 3.2 && 352 ± 43 & 2735 ± 109 & 199 ± 00\\
                 & Chemformer & 91.0 ± 2.9 && 381 ± 50 & 30209 ± 242 & 75.2 ± 4.2\\

\bottomrule
\end{tabular}
\end{table*}

\begin{table*}[hbt]
\small
\centering
\caption{\label{supptab:caspyrus_ms_subsample_500} Multi-step synthesis planning metrics for a subsample size of 500
Caspyrus10k molecules. The performance is measured for each single-step model and dataset by randomly subsampling 1000 times with the subsample size (sampling without replacement). For each subsample, the same molecules are used across single-step models and datasets.
}
\begin{tabular}{@{}rrrcrrr@{}}\toprule
& & \multicolumn{1}{c}{Overall} & \phantom{}& \multicolumn{3}{c}{Average per Molecule} \\
\cmidrule{3-3} \cmidrule{5-7}
\textbf{Training Dataset} & \textbf{Model} & \textbf{Success Rate (\%)} && \textbf{Solved Routes} & \textbf{Search Time (s)} & \textbf{Model Calls} \\ \midrule
\multirow{4}{*}{\begin{tabular}[r]{@{}l@{}}USPTO-50k\end{tabular}}         
                 & AZF        & 41.1 ± 2.1 && 36.2 ± 3.7 & 159 ± 0 & 198 ± 0\\
                 & LocalRetro & 74.1 ± 1.8 && 124 ± 7 & 160 ± 2 & 200 ± 0\\
                 & Chemformer & 62.5 ± 2.1 && 7.38 ± 0.66 & 19028 ± 337 & 176 ± 2 \\
                 & MHNreact   & 51.1 ± 2.2 && 38.6 ± 3.4 & 28956 ± 4 & 99.4 ± 1.1 \\ \hline
\multirow{4}{*}{\begin{tabular}[r]{@{}l@{}}USPTO-PaRoutes-1M\end{tabular}}         
                 & AZF        & 66.4 ± 2.0 && 83.6 ± 5.8 & 162 ± 0 & 199 ± 0\\
                 & LocalRetro & 86.1 ± 1.5 && 325 ± 18 & 1216 ± 21 & 200 ± 0\\
                 & Chemformer & 94.2 ± 1.0 && 463 ± 26 & 28811 ± 41 & 147 ± 1\\
                 & MHNreact   & 64.7 ± 2.1 && 215 ± 15 & 28838 ± 10 & 169 ± 0\\ \hline
\multirow{4}{*}{\begin{tabular}[r]{@{}l@{}}AZ-1M\end{tabular}}             
                 & AZF        & 73.7 ± 1.9 && 124 ± 7 & 168 ± 1 & 199 ± 0\\
                 & LocalRetro & 88.1 ± 1.4 && 322 ± 16 & 464 ± 15 & 199 ± 0\\
                 & Chemformer & 94.5 ± 1.0 && 358 ± 19 & 29108 ± 29 & 107 ± 1\\
                 & MHNreact   & 56.0 ± 2.2 && 77.2 ± 7.1 & 29116 ± 15 & 64.6 ± 1.4\\ \hline
\multirow{3}{*}{\begin{tabular}[r]{@{}l@{}}AZ-18M\end{tabular}}            
                 & AZF        & 76.4 ± 1.8 && 154 ± 9 & 153 ± 2 & 199 ± 0\\
                 & LocalRetro & 87.3 ± 1.4 && 351 ± 19 & 2732 ± 48 & 199 ± 0\\
                 & Chemformer & 91.0 ± 1.2 && 380 ± 22 & 30212 ± 110 & 75.1 ± 1.8\\
\bottomrule
\end{tabular}
\end{table*}

\begin{table*}[hbt]
\small
\centering
\caption{\label{supptab:caspyrus_ms_subsample_1000} Multi-step synthesis planning metrics for a subsample size of 1,000
Caspyrus10k molecules. The performance is measured for each single-step model and dataset by randomly subsampling 1000 times with the subsample size (sampling without replacement). For each subsample, the same molecules are used across single-step models and datasets.
}
\begin{tabular}{@{}rrrcrrr@{}}\toprule
& & \multicolumn{1}{c}{Overall} & \phantom{}& \multicolumn{3}{c}{Average per Molecule} \\
\cmidrule{3-3} \cmidrule{5-7}
\textbf{Training Dataset} & \textbf{Model} & \textbf{Success Rate (\%)} && \textbf{Solved Routes} & \textbf{Search Time (s)} & \textbf{Model Calls} \\ \midrule
\multirow{4}{*}{\begin{tabular}[r]{@{}l@{}}USPTO-50k\end{tabular}}         
                 & AZF        & 41.1 ± 1.4 && 36.1 ± 2.6 & 159 ± 0 & 198 ± 0\\
                 & LocalRetro & 74.0 ± 1.3 && 124 ± 5 & 160 ± 1 & 200 ± 0\\
                 & Chemformer & 62.4 ± 1.4 && 7.35 ± 0.47 & 19061 ± 245 & 176 ± 1 \\
                 & MHNreact   & 50.9 ± 1.5 && 38.5 ± 2.3 & 28956 ± 3 & 99.4 ± 0.7 \\ \hline
\multirow{4}{*}{\begin{tabular}[r]{@{}l@{}}USPTO-PaRoutes-1M\end{tabular}}         
                 & AZF        & 66.3 ± 1.5 && 83.5 ± 4.1 & 162 ± 0 & 199 ± 0\\
                 & LocalRetro & 86.0 ± 1.1 && 324 ± 13 & 1218 ± 15 & 200 ± 0\\
                 & Chemformer & 94.1 ± 0.7 && 463 ± 18 & 28811 ± 29 & 147 ± 0\\
                 & MHNreact   & 64.6 ± 1.5 && 214 ± 11 & 28839 ± 7 & 169 ± 0\\ \hline
\multirow{4}{*}{\begin{tabular}[r]{@{}l@{}}AZ-1M\end{tabular}}             
                 & AZF        & 73.5 ± 1.4 && 124 ± 5 & 168 ± 1 & 199 ± 0\\
                 & LocalRetro & 88.0 ± 1.0 && 321 ± 11 & 465 ± 10 & 199 ± 0\\
                 & Chemformer & 94.4 ± 0.7 && 358 ± 13 & 29108 ± 20 & 107 ± 1\\
                 & MHNreact   & 56.0 ± 1.5 && 76.9 ± 5.1 & 29115 ± 10 & 64.6 ± 0.9\\ \hline
\multirow{3}{*}{\begin{tabular}[r]{@{}l@{}}AZ-18M\end{tabular}}            
                 & AZF        & 76.2 ± 1.3 && 154 ± 6 & 153 ± 1 & 199 ± 0\\
                 & LocalRetro & 87.3 ± 1.0 && 350 ± 13 & 2737 ± 33 & 199 ± 0\\
                 & Chemformer & 90.9 ± 0.9 && 381 ± 14 & 30210 ± 79 & 75.1 ± 1.3\\
\bottomrule
\end{tabular}
\end{table*}

\begin{table*}[hbt]
\small
\centering
\caption{\label{supptab:caspyrus_ms_subsample_5000} Multi-step synthesis planning metrics for a subsample size of 5,000
Caspyrus10k molecules. The performance is measured for each single-step model and dataset by randomly subsampling 1000 times with the subsample size (sampling without replacement). For each subsample, the same molecules are used across single-step models and datasets.
}
\begin{tabular}{@{}rrrcrrr@{}}\toprule
& & \multicolumn{1}{c}{Overall} & \phantom{}& \multicolumn{3}{c}{Average per Molecule} \\
\cmidrule{3-3} \cmidrule{5-7}
\textbf{Training Dataset} & \textbf{Model} & \textbf{Success Rate (\%)} && \textbf{Solved Routes} & \textbf{Search Time (s)} & \textbf{Model Calls} \\ \midrule
\multirow{4}{*}{\begin{tabular}[r]{@{}l@{}}USPTO-50k\end{tabular}}         
                 & AZF        & 41.1 ± 0.5 && 36.0 ± 0.9 & 159 ± 0 & 198 ± 0\\
                 & LocalRetro & 74.0 ± 0.4 && 124 ± 1 & 160 ± 0 & 200 ± 0\\
                 & Chemformer & 62.3 ± 0.5 && 7.37 ± 0.16 & 19053 ± 79 & 176 ± 0 \\
                 & MHNreact   & 50.9 ± 0.5 && 38.4 ± 0.8 & 28956 ± 1 & 99.3 ± 0.2 \\ \hline
\multirow{4}{*}{\begin{tabular}[r]{@{}l@{}}USPTO-PaRoutes-1M\end{tabular}}         
                 & AZF        & 66.3 ± 0.5 && 83.4 ± 1.4 & 162 ± 0 & 199 ± 0\\
                 & LocalRetro & 86.0 ± 0.3 && 324 ± 4 & 1218 ± 4 & 200 ± 0\\
                 & Chemformer & 94.1 ± 0.2 && 463 ± 6 & 28810 ± 10 & 147 ± 0\\
                 & MHNreact   & 64.6 ± 0.5 && 214 ± 3 & 28838 ± 2 & 169 ± 0\\ \hline
\multirow{4}{*}{\begin{tabular}[r]{@{}l@{}}AZ-1M\end{tabular}}             
                 & AZF        & 73.5 ± 0.4 && 124 ± 1 & 168 ± 0 & 199 ± 0\\
                 & LocalRetro & 88.0 ± 0.3 && 321 ± 3 & 465 ± 3 & 199 ± 0\\
                 & Chemformer & 94.4 ± 0.2 && 358 ± 4 & 29109 ± 7 & 107 ± 0\\
                 & MHNreact   & 55.9 ± 0.5 && 77.0 ± 1.7 & 29115 ± 3 & 64.6 ± 0.3\\ \hline
\multirow{3}{*}{\begin{tabular}[r]{@{}l@{}}AZ-18M\end{tabular}}            
                 & AZF        & 76.2 ± 0.4 && 154 ± 2 & 153 ± 0 & 199 ± 0\\
                 & LocalRetro & 87.3 ± 0.3 && 350 ± 4 & 2737 ± 10 & 199 ± 0\\
                 & Chemformer & 90.9 ± 0.3 && 380 ± 4 & 30212 ± 26 & 75.1 ± 0.4\\
\bottomrule
\end{tabular}
\end{table*}


\begin{table*}[hbt]
\small
\centering
\caption{\label{supptab:caspyrus_ms_selected_1000_mol_subsample} Multi-step synthesis planning metrics for the provided randomly selected subsample of 1,000
Caspyrus10k molecules.
}
\begin{tabular}{@{}rrrcrrr@{}}\toprule
& & \multicolumn{1}{c}{Overall} & \phantom{}& \multicolumn{3}{c}{Average per Molecule} \\
\cmidrule{3-3} \cmidrule{5-7}
\textbf{Training Dataset} & \textbf{Model} & \textbf{Success Rate (\%)} && \textbf{Solved Routes} & \textbf{Search Time (s)} & \textbf{Model Calls} \\ \midrule
\multirow{4}{*}{\begin{tabular}[r]{@{}l@{}}USPTO-50k\end{tabular}}         
                 & AZF        & 40.7 && 37.5 & 159 & 198 \\
                 & LocalRetro & 73.6 && 125 & 163 & 200 \\
                 & Chemformer & 62.9 && 7.09 & 19269 & 176 \\
                 & MHNreact   & 50.0 && 39.3 & 28955 & 99.0 \\ \hline
\multirow{4}{*}{\begin{tabular}[r]{@{}l@{}}USPTO-PaRoutes-1M\end{tabular}}         
                 & AZF        & 67.4 && 87.7 & 162 & 199 \\
                 & LocalRetro & 85.6 && 327 & 1231 & 200 \\
                 & Chemformer & 94.1 && 448 & 28752 & 146 \\
                 & MHNreact   & 63.6 && 204 & 28845 & 168 \\ \hline
\multirow{4}{*}{\begin{tabular}[r]{@{}l@{}}AZ-1M\end{tabular}}            
                 & AZF        & 74.8 && 126 & 169 & 199 \\
                 & LocalRetro & 88.6 && 325 & 466 & 200 \\
                 & Chemformer & 94.2 && 382 & 29087 & 108 \\
                 & MHNreact   & 54.5 && 75.8 & 29113 & 65.1 \\ \hline
\multirow{3}{*}{\begin{tabular}[r]{@{}l@{}}AZ-18M\end{tabular}}            
                 & AZF        & 77.4 && 156 & 154 & 199 \\
                 & LocalRetro & 87.5 && 351 & 2783 & 200 \\
                 & Chemformer & 90.7 && 391 & 30127 & 76.1 \\
\bottomrule
\end{tabular}
\end{table*}


\clearpage
\subsubsection{PaRoutes}

\begin{table*}[hbt]
\small
\centering
\caption{\label{supptab:paroutes_performance} Multi-step synthesis planning route accuracy (\subref{supptab:paroutes_route_accuracy}) and building block accuracy (\subref{supptab:paroutes_building_block_accuracy}) on PaRoutes gold-standard synthesis routes with
different single-step models trained on USPTO-PaRoutes-1M.
}

\begin{subtable}[h]{\textwidth}
\centering
\caption{\label{supptab:paroutes_route_accuracy} Route Accuracy} 
    \begin{tabular}{@{}rrrrrrrr@{}}\toprule
\textbf{Training Dataset} & \textbf{Model} & \textbf{Top-1} & \textbf{Top-3} & \textbf{Top-5} & \textbf{Top-10} & \textbf{Top-50}  \\ \midrule
\multirow{4}{*}{\begin{tabular}[r]{@{}l@{}}USPTO-PaRoutes-1M\end{tabular}} & AZF & 23.7 & 48.5 & 56.5 & 60.7 & 61.8 \\
                 & LocalRetro & 3.72 &  9.92 & 13.8 & 20.2 & 36.0 \\
                 & Chemformer & 1.9 & 5.8 & 9.4 & 13.8 & 26.5 \\
                 & MHNreact   &  4.2 & 11.3 & 16.0 & 22.9 & 39.7 \\ 
\bottomrule
\end{tabular}
\end{subtable}

\vspace*{0.5cm}

\begin{subtable}[h]{\textwidth}
\centering
\caption{\label{supptab:paroutes_building_block_accuracy} Building Block Accuracy} 
    \begin{tabular}{@{}rrrrrrrr@{}}\toprule
\textbf{Training Dataset} & \textbf{Model} & \textbf{Top-1} & \textbf{Top-3} & \textbf{Top-5} & \textbf{Top-10} & \textbf{Top-50}  \\ \midrule
\multirow{4}{*}{\begin{tabular}[r]{@{}l@{}}USPTO-PaRoutes-1M\end{tabular}} & AZF & 45.3 & 64.1 & 71.2 & 75.2 & 76.0 \\
                 & LocalRetro & 16.4 & 28.3 & 34.7 & 43.8 & 62.6 \\
                 & Chemformer & 9.8 & 20.3 & 25.8 & 33.9 & 49.5\\
                 & MHNreact   &  15.6 & 26.9 & 33.2 & 41.3 & 57.1\\ 
\bottomrule
\end{tabular}
\end{subtable}

\end{table*}

\begin{figure}[h]
    \centering
    \includegraphics[width=0.9\textwidth]{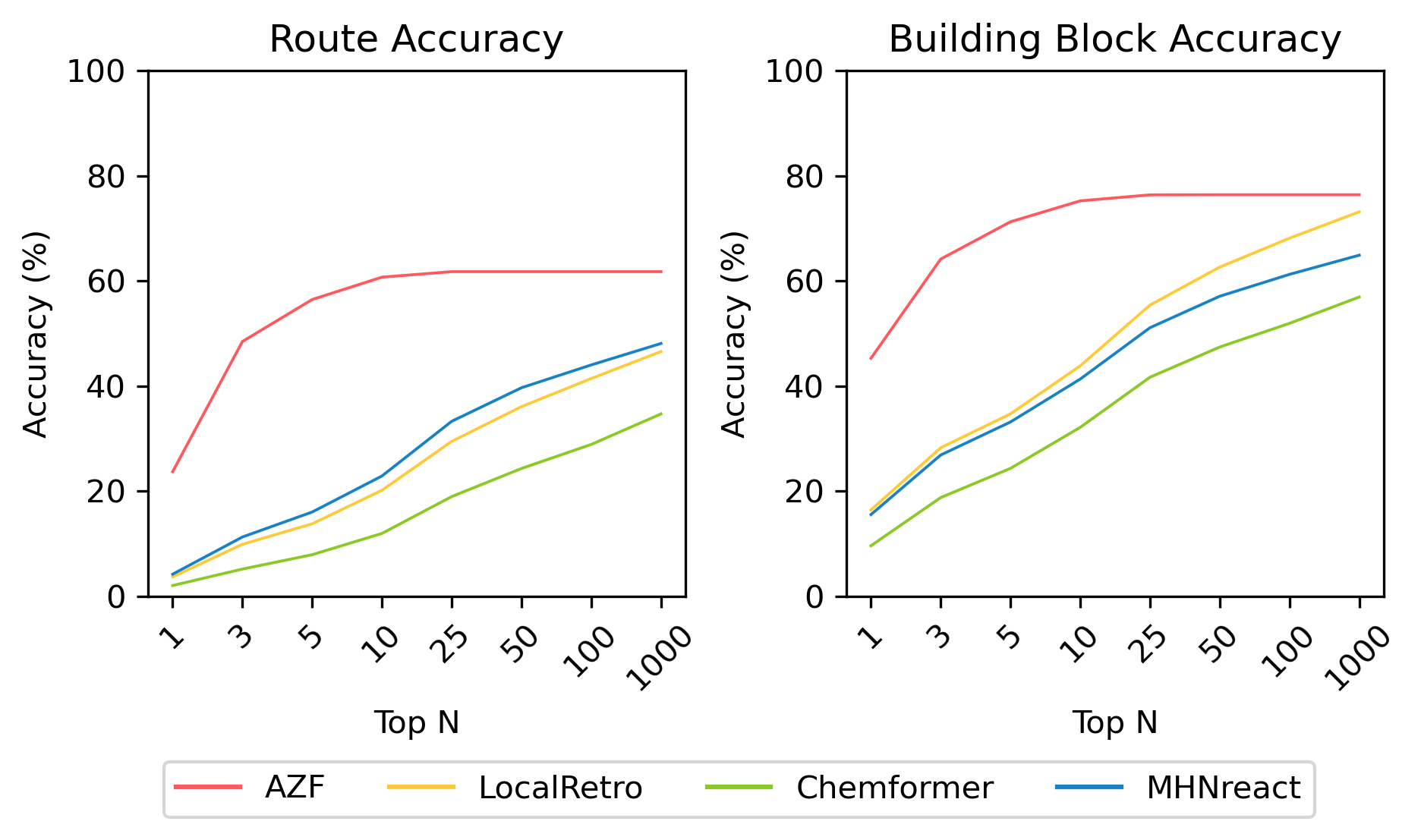} 
    \caption{Multi-step synthesis planning accuracy up to top-1000 on PaRoutes gold-standard synthesis routes with different single-step models trained on USPTO-PaRoutes-1M. Route accuracy measures the ability to recover the correct synthesis route within top-n, whereas building block accuracy measures the ability to recover the correct building blocks while not considering reactions and intermediates.}
    \label{suppfig:paroutes_accuracy_1k}
\end{figure}

\begin{figure}[h]
    \centering
    \includegraphics[width=0.6\textwidth]{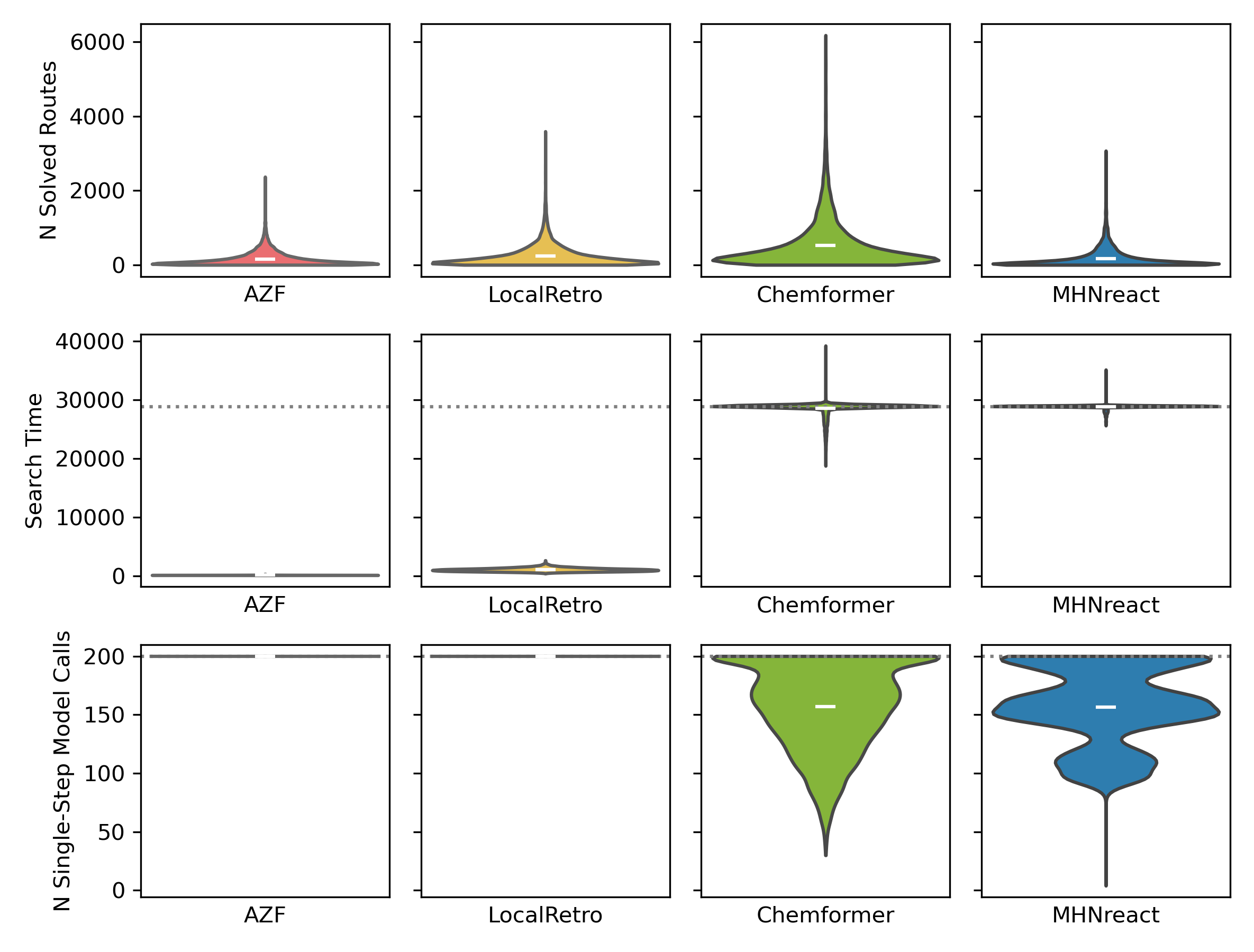} 
    \caption{Distributions of solved routes, search time and single-step model calls for synthesis planning results of single-step models trained on USPTO-PaRoutes-1M and evaluated on PaRoutes. The dashed line indicates the respective limits set in algorithm search settings. The white line indicates the mean across all molecules for the shown model-training set combination.}
    \label{suppfig:ms_metrics_paroutes}
\end{figure}

\begin{figure}[h]
    \centering
    \includegraphics[width=0.8\textwidth]{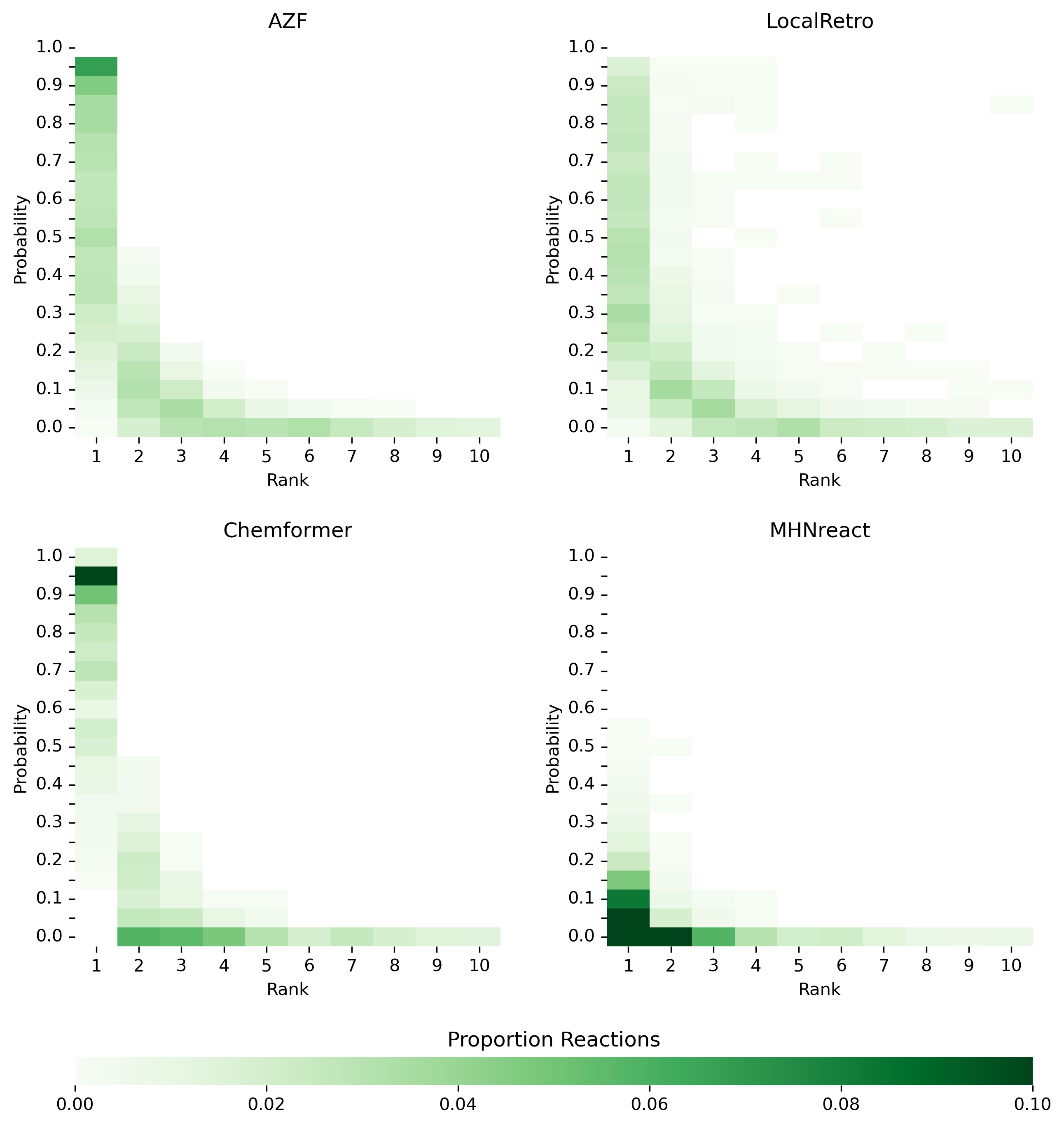} 
    \caption{Single-step model prior and rank distributions of reactions from the correctly predicted PaRoutes synthesis routes. Reactions are extracted from the top-10 predicted routes for each single-step retrosynthesis model trained on USPTO-PaRoutes-1M.}
    \label{suppfig:ms_priors_accurate}
\end{figure}

\end{document}